\documentclass[10pt,journal]{IEEEtran}
\ifCLASSINFOpdf
\else
\fi
\usepackage{url}
\usepackage{graphicx}
\usepackage{amssymb}
\usepackage{amsfonts}
\usepackage{mathrsfs}
\usepackage{amsmath}
\usepackage{algorithm}
\usepackage{algorithmic}
\usepackage{multirow}
\usepackage{booktabs}
\usepackage[table]{xcolor}
\usepackage{cite}
\usepackage{setspace}
\usepackage{bm}
\usepackage{multirow}
\usepackage{makecell}
\allowdisplaybreaks[2] 
\definecolor{mygray}{gray}{.9}
\definecolor{babyblueeyes}{rgb}{0.7, 0.8, 1}

\renewcommand{\arraystretch}{1}

\DeclareMathOperator*{\argmin}{argmin}

\begin{document}
%
\title{Hyperspectral Image Restoration via Total Variation Regularized Low-rank Tensor Decomposition}

\author{ 
         Yao Wang,~
         Jiangjun Peng,~
         Qian Zhao,~Deyu Meng,~\IEEEmembership{Member,~IEEE,}
        Yee Leung
        and Xi-Le Zhao
\thanks{Yao Wang, Jiangjun Peng, Qian Zhao and Deyu Meng are with School of Mathematics and Statistics, Xi'an Jiaotong University, Xi'an 710049, China. E-mails: yao.s.wang@gmail.com, andrew.pengjj@gmail.com, timmy.zhaoqian@gmail.com, dymeng@mail.xjtu.edu.cn.}
\thanks{Yee Leung is with the Institute of Future Cities, Chinese University of Hongkong, e-mail: yeeleung@cuhk.edu.hk.}
\thanks{Xi-Le Zhao is with School of Mathematical Sciences, University of Electronic Science and Technology of China, Chengdu 610054, e-mail: xlzhao122003@163.com.}
\thanks{Qian Zhao is the corresponding author.}

}
\maketitle

\begin{abstract}
Hyperspectral images (HSIs) are often corrupted by a mixture of several types of noise during the acquisition process, e.g., Gaussian noise, impulse noise, dead lines, stripes, and many others. Such complex noise could degrade the quality of the acquired HSIs, limiting the precision of the subsequent processing. In this paper, we present a novel tensor-based HSI restoration approach by fully identifying the intrinsic structures of the clean HSI part and the mixed noise part respectively. Specifically, for the clean HSI part, we use tensor Tucker decomposition to describe the global correlation among all bands, and an anisotropic  spatial-spectral total variation (SSTV) regularization to characterize the piecewise smooth structure in both spatial and spectral domains. For the mixed noise part, we adopt the $\ell_1$ norm regularization to detect the sparse noise, including stripes, impulse noise, and dead pixels. Despite that TV regulariztion has the ability of removing Gaussian noise, the Frobenius norm term is further used to model heavy Gaussian noise for some real-world scenarios. Then, we develop an efficient algorithm for solving the resulting optimization problem by using the augmented Lagrange multiplier (ALM) method.  Finally, extensive experiments on simulated and real-world noise HSIs are carried out to demonstrate the superiority of the proposed method over the existing state-of-the-art ones.

\end{abstract}

\begin{IEEEkeywords}
Hyperspectral image (HSI), mixed noise,  low rank tensor decomposition, total variation (TV). 
\end{IEEEkeywords}


%
\IEEEpeerreviewmaketitle

\section{Introduction}
Hyperspectral imaging employs an imaging spectrometer to collect hundreds of spectral bands ranging from ultraviolet to infrared wavelengths for the same area on the surface of the Earth. It has a wide range of applications including environmental monitoring, military surveillance,  mineral exploration, among numerous others \cite{Goetz2009remote}, \cite{willett2014sparsity}. Due to  various factors,  e.g., thermal  electronics, dark current,  and  stochastic  error  of  photo-counting in imaging process, hyperspectral images (HSIs) are inevitably corrupted by severe noise during the acquisition process. This greatly degrades the visual quality of the HSIs and further affects the precision of previous listed applications. Hence, the task of removing the noise in hyperspectral imagery is a valuable research topic and has received much attention in the past decades. 

A natural way for restorazing HSIs is to regard each band as a gray-level image and then apply traditional 2-D or 1-D denoising methods to remove noise band-by-band.  See, e.g., \cite{green1988transformation, elad2006image, dabov2007image, mairal2009non}. However, this kind of method ignores the correlations among all the spectral bands or spatial pixels, and thus usually could not provide satisfactory results. Aiming at taking account of such correlations, a variety of studies have been conducted in the literature. For example, Othman and Qian \cite{othman2006noise}  proposed a hybrid spatial-sepectral wavelet shrinkage method to take advantage of dissimilarity of the signal regularity in both the spatial and spectral domains of the HSIs. Zhong and Wang \cite{zhong2013multiple} designed a multiple spectral-band conditional random field method, which can simultaneously model and utilize the spatial and spectral dependences in a unified probabilistic framework. In \cite{yuan2012hyperspectral}, an efficient HSI denoising procedure is proposed by designing a spectral-spatial adaptive total variation model,  where the spectral noise differences and spatial information differences were both considered. Additionally, several advanced techniques in traditional image processing, including the nonlocal similarity \cite{qian2013nonlocal}, wavelet shrinkage \cite{chen2011denoising}, anisotropic diffusion \cite{wang2010anisotropic},  have been adopted for HSI restoration in the recent years.  

\begin{figure}[!]
\centering
\includegraphics[scale=0.65]{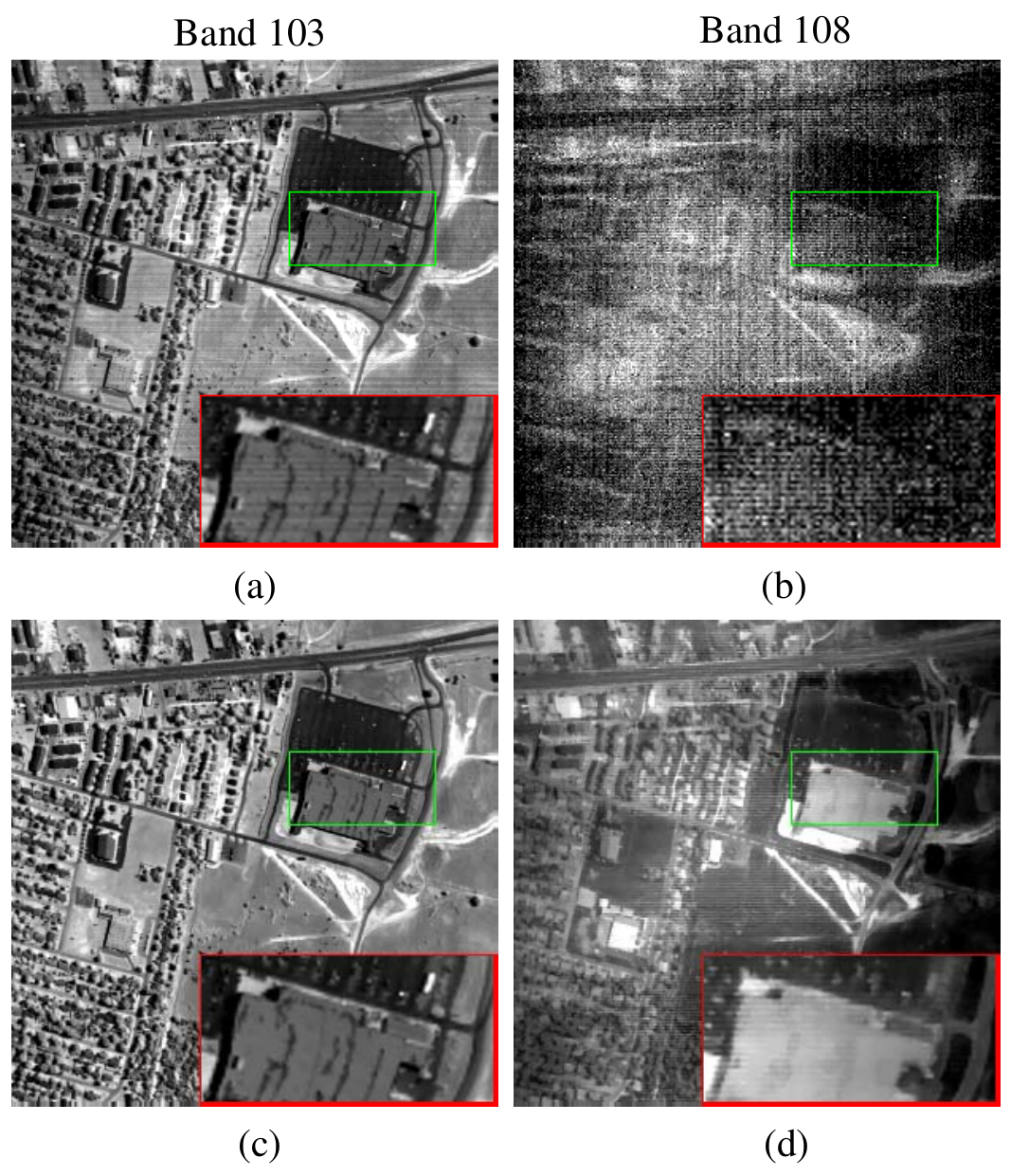}
\caption{A typical restoration instance on HYDICE urban data: (a-b) the original band; (c-d) denoised results by LRTV. }
\label{fig_urban}
\end{figure}  

Actually, all the aforementioned methods  are designed to remove only one or two types of noise, i.e., Gaussian noise, impulse noise, or hybrid Gaussian-impulse noise. In real-world scenarios, however, HSIs are usually corrupted by a combination of several different types of noise during the acquisition process, e.g., Gaussian noise, impulse noise, dead lines, stripes, and many others. Although several methods based on low-rank matrix modeling have been proposed for removing mixed noise in HSIs,  e.g., \cite{zhang2014hyperspectral, song2014hyperspectral, he2015hyperspectral, wang2016denoising, he2016total}, modeling a HSI cube as a matrix is not able to utilize the finer spatial-and-spectral information,  leading to suboptimal restoration results in some heavy noisy situations.  See Fig. 1 for an instance. It can be easily to observe that, for the HSI band polluted by severe noise (e.g., band 108), the state-of-the-art low-rank matrix method, e.g., LRTV proposed by \cite{he2016total}, could not provide satisfactory restoration as for moderately noisy HSI band (e.g., band 103). 

Recently, a number of studies, both practically and theoretically,  have demonstrated the advantages of direct tensor modeling techniques over matricization techniques in dealing with high-order tensor data. See, e.g., \cite{Liu2013tensor, Yuan2014tensor, Cao2015tensor, Anima2016tensor, Lu2016tensor} among others. Motivated by such studies, we propose in this paper a novel mixed noise removal approach using direct tensor modeling techniques to fully exploit the spatial-and-spectral priors underlying the clean HSI part and characterize the intrinsic structures of the heavy noise part. To highlight our contributions, we shall go over related work on HSI restoration using different low-rank modeling techniques and contrast our innovations with the existing literature.


\subsection{Related Work}
In the past few years,  various approaches based on low-rank matrix approximation (LRMA) have  been proposed  for HSI restoration, and can be represented as state-of-the-art techniques. Motivated by the idea of robust principle component analysis (RPCA) \cite{Candes2011rpca}, Zhang et al. \cite{zhang2014hyperspectral} explored the low-rank property by lexicographically ordering a patch of the HSI into a 2-D matrix and modeled the non-Gaussian noise (including impulse noise, dead lines, and stripes) as a sparse part. The so-called ``Go Decomposition'' (GoDec) \cite{zhou2011godec} algorithm was used to enforce rank $r$ and cardinality $k$ constraints for the low-rank and sparse parts, respectively. Considering that the noise intensity in different bands is different, a noise-adjusted iterative LRMA method on the basis of patchwise randomized singular value decomposition was proposed in \cite{he2015hyperspectral}. Encouraged by the powerfulness of total variation (TV) regularization in various image restoration tasks, He et al. \cite{he2016total} integrated the band-by-band TV regularization into a rank-constrained RPCA model to explore the spatial piecewise smoothness of the HSI, and as a result enhanced the capability of the LRMA technique for HSI restoration. Similarly, Wu et al. \cite{wu2016total} proposed a HSI mixed denoising model by combining the band-by-band TV regularization with the widely used weighted nuclear norm minimization (WNMM) \cite{gu2014weighted, peng2014reweighted}. Aiming at giving better approximation to the low-rank assumption of HSI data, a weighted Schatten $p$-norm regularization was introduced
by Xie et al. \cite{xie2016hyperspectral} into the LRMA framework. To exploit both the local similarity within a HSI patch and the nonlocal similarity across patches in a group simultaneously, a novel group low-rank representation model was consider in \cite{wang2016denoising}. 
As stated before, though this kind of LRMA methodology has
been an increasingly useful technique in HSI restoration, it fails to fully exploit the prior knowledge on the intrinsic structures of HSI cube after vectorizing the HSI bands. 

Despite the efficiency of  tensor methods on removing Gaussian noise or  hybrid Gaussian-impulse noise in HSIs, e.g., \cite{renard2008denoising, karami2011noise, liu2012denoising, guo2013hyperspectral, peng2014decomposable, xie2016multispectral}, to the best of our knowledge, only two studies  \cite{li2015hyperspectral, wu2017structure} based on tensor techniques have been conducted to remove mixed noise in HSIs. Specifically, in  \cite{li2015hyperspectral}, the tensor nuclear norm introduced by Liu et al. \cite{Liu2013tensor} was used to  acquire low-rank property of HSI cube, and the mixed $\ell_2/\ell_1$ norm was used to impose sparsity property of the outliers and non-Gaussian noise; the work \cite{wu2017structure} integrated the  structure tensor TV \cite{lefkimmiatis2015structure} into the WNNM model,  and demonstrated outperformance over band-by-band TV-regularized WNMM method \cite{wu2016total}, because of the utilization of finer spatial structure information.

Basically, our work is related to the aforementioned works, however,  there exits significant differences between our work and other ones. Fig. 2 illustrates the frameworks of the popular matrix modeling idea especially LRMA and our direct tensor modeling idea in dealing with noisy HSI cubes. It can be seen that,  the matricization techniques  should  preliminarily vectorize all HSI bands at the cost of losing spatial structures of the HSI cube, whereas direct tensor modeling techniques could deliver more faithful underlying information of the HSI cube without destroying the spatial structures.  Despite their connection and similarity, compared with the work \cite{li2015hyperspectral}, we adopt an anisotropic spatial-spectral TV to further exploit the local  piecewise smoothness in both spatial and spectral domains; compared with the work \cite{wu2017structure}, low-rank Tucker decomposition is used to further characterize the spatial correlation in each HSI band, and the proposed spatial-spectral TV is more simple than the structure tensor TV. Additionally, the use of Frobenius norm term for dealing with very large Gaussian noise was not consider in \cite{li2015hyperspectral,wu2017structure}. A detailed motivation of our work can be found in Section III. 

\begin{figure}[!]
\centering
\includegraphics[scale=0.29]{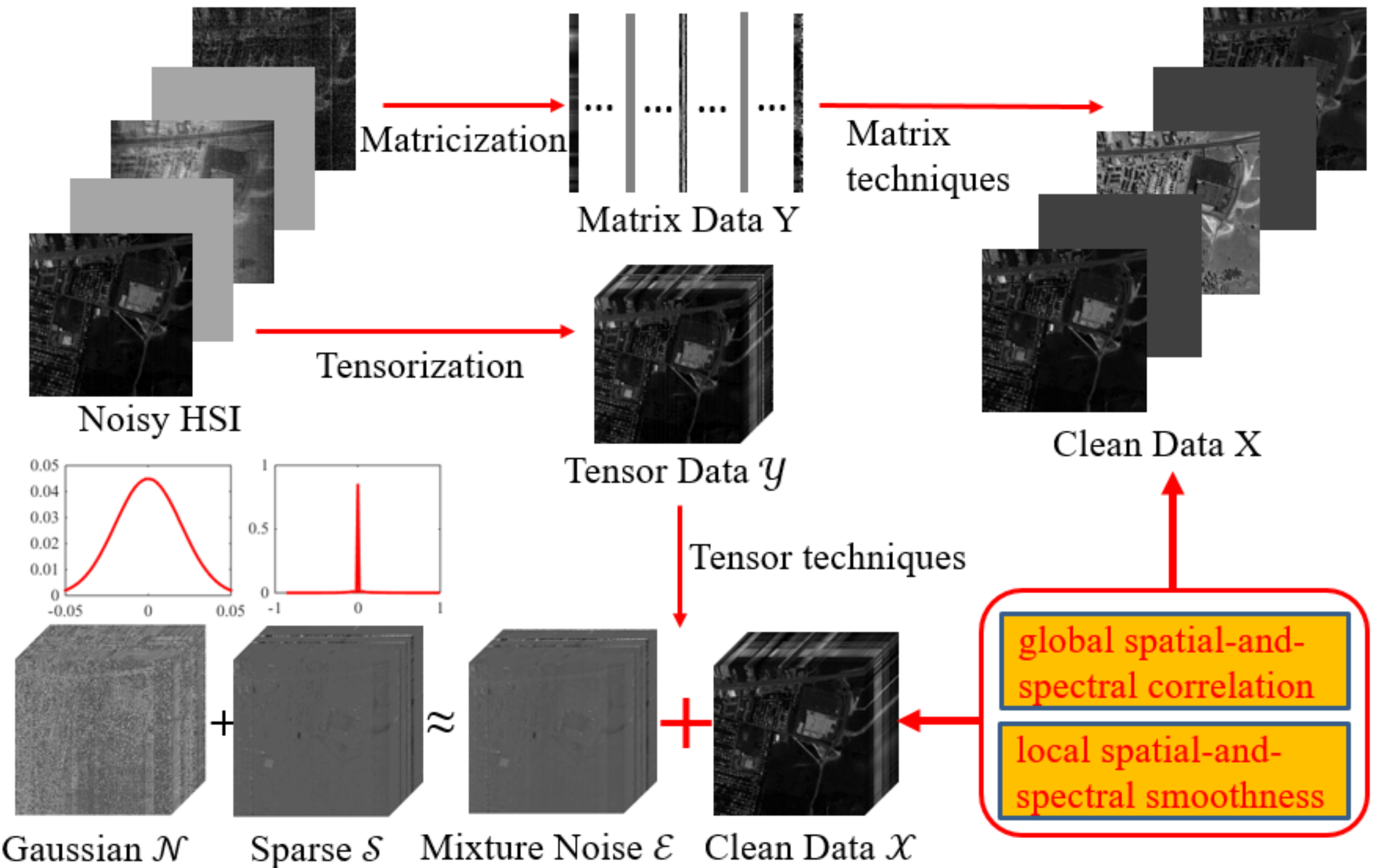}
\caption{Schematic diagrams of popular matrix procedure and our tensor procedure for HSI restoration. }
\label{fig_urban}
\end{figure}  
\subsection{Our contributions}
In this paper, we mainly focus on HSI mixed noise removal. The main contributions of this paper are summarized as follows.
\begin{enumerate}
\item[1)] The low-rank Tucker decomposition is applied to separate the clean HSI from
the raw observation corrupted by complex nose, which can be fully capable of exploiting  the global spatial-and-spectral correlation among the HSI bands. The classical HOOI algorithm is used to achieve such Tucker decomposition efficiently, without bringing extra computational burden.

\item[2)] A designed spatial-spectral total variation ({SSTV}) regularizer is incorporated into the low-rank Tucker decomposition framework. It is shown that the SSTV regularization has the ability of  enhancing the spatial information and thus preserves the spectral signatures of HSIs very well. 

\item[3)]  The well-known augmented Lagrange multiplier (ALM) method is used for solving the TV regularized low rank tensor decomposition model. And extensive studies on simulated and real data  experiments are carried out to illustrate that the proposed method clearly improves the HSI restoration results over some other popular techniques, in terms of both the quantitative evaluation and the visual comparison.
\end{enumerate}

 \begin{table}[t]
\renewcommand\arraystretch{1.3}
\centering
\caption{Notations}
\begin{tabular}{lp{0.05cm}p{5cm}} 
\toprule[1pt]
Notations & & Explanations \\
\hline
 $\mathcal{X}$, $\mathbf{X}$, $\mathbf{x}$, $x$   &    & Tensor, matrix, vector, scalar.   \\

$\mathbf{X}_{(n)}$ or $\mathcal{X}_{(n)}$ & &Mode-$n$ matricization of tensor $\mathcal{X}$ $\in$ $\Re^{I_1\times I_2\times,\cdots,\times I_N}$, obtained by arranging the mode-$n$ fibers as the columns of the resulting matrix of size $\Re^{I_n \times \prod_{k \neq n} I_k}$. \\
$(r_1,r_2,\cdots,r_N)$  & & Multi-linear rank, where $r_n = \text{rank}(\textbf{X}_{(n)})$, $n = 1,2,\cdots,N.$ \\
$\langle \mathcal{X}, \mathcal{Y} \rangle$     &     & Inner product of tensor $\mathcal{X}$ and $\mathcal{Y}$. \\
$\| \mathcal{X} \|_{F}$                        &     & Frobenius norm of tensor $\mathcal{X}$. \\
$\mathcal{Y}=\mathcal{X} \times_n \textbf{U} $ &    & Mode-$n$ multiplication of $\mathcal{X}$ and $\textbf{U}$ with the matrix representation $\textbf{Y}_{(n)}= \textbf{U} \textbf{X}_{(n)}$. \\
\bottomrule[1pt]
\end{tabular}
\label{fuhao}
\vspace{-2mm}
\end{table}

\subsection{Organization of The Paper}
This paper is organized as follows. Section II introduces some notations and preliminaries of tensors, which will be used for presenting our procedure. In Section III, the anisotropic SSTV regularized low-rank tensor decomposition model and its motivations are introduced. We then develop an efficient ALM algorithm for solving the proposed model.  In Section IV, extensive experiments on both simulated and real datasets are carried out to illustrate the merits of our model. Finally, we conclude this paper with some discussions on future research in Section V.
\section{Notation and Preliminaries}
It is known that a tensor can be seen as a multi-index numerical array, and its order is defined as the number of its modes or dimensions. A real-valued tensor of order $N$ is denoted by $\mathcal{X}\in\mathbb{R}^{I_1\times I_2\times\ldots\times I_N}$ and its entries by $x_{i_1,i_2,\ldots,i_N}$. We then can consider an $N\times 1$ vector $x$  as a tensor of order one, and an $N\times M$ matrix $\mathbf{X}$ as a tensor of order two. 
Following \cite{K2009Tensor}, we shall provide a brief introduction on tensor algebra. 

The inner product of two same-sized tensors $\mathcal{X}, \mathcal{Y}$ is defined as 
$\langle \mathcal{X}, \mathcal{Y}\rangle :=\sum\limits_{i_1,i_2,\ldots,i_N}x_{i_1,i_2,\ldots,i_N}\cdot y_{i_1,i_2,\ldots,i_N}$. Then the corresponding Frobenius norm is defined as  $\|\mathcal{X}\|_F=\sqrt{\langle\mathcal{X}, \mathcal{X}\rangle}$. The so-called mode-$n$ matricization of a tensor $\mathcal{X}$ is denoted as $\mathbf{X}_{(n)}$,  where the tensor element $(i_1,i_2, \ldots, i_N)$ maps to the matrix element  $(i_n, j)$ satisfying $j=1+\sum_{k=1,k\neq n}^N(i_k-1)J_k$ with $J_k=\prod_{m=1,m\neq n}^{k-1}I_m$. The
mode-$n$ multiplication of a tensor $\mathcal{X}$ with a matrix $\mathbf{U}$ is denoted by $\mathcal{X}\times_n\mathbf{U}$ and elementwise,  we have $(\mathcal{X}\times_n\mathbf{U})_{i_1,\ldots,i_{n-1}ji_{n+1}\ldots,i_N}=\sum_{i_n}x_{i_1,i_2,\ldots,i_N}\cdot u_{j,i_n}.$  The multi-linear rank is defined as an array $(r_1, r_2, \ldots, r_N)$ where $r_n=\mbox{rank}(\mathbf{X}_{(n)})$, $n=1,2,\ldots, N$. 

The tensor notations used in this work are summarized in Table I. Interested readers are  referred to \cite{K2009Tensor} for a more detailed introduction. 
\section{HSI Restoration Via TV Regularized Low-rank Tensor Decomposition}
\subsection{Motivation} 
In many real situations, the observed HSI data are contaminated by a mixture of several different kinds of noise \cite{zhang2014hyperspectral, he2015hyperspectral, he2016total}. As a result, a noise HSI cube denoted by a three order tensor $\mathcal{X}:=\{\mathbf{X}^{1}$,$\textbf{X}^2,\ldots,\mathbf{X}^{B}\}$, where each matrix $\textbf{X}^{i} \in \mathbb{R}^{h\times w}(i=1,2,\ldots,B)$ represents $i$-th band, with the height $h$  and width $w$, and $B$ denotes the number of bands, can be described as
\begin{equation}
\mathcal{Y}=\mathcal{X}+\mathcal{E},
\end{equation}
where $\mathcal{X}$ and $\mathcal{E}$ are with the same size of $\mathcal{Y}$, which represent the clean HSI cube and the mixed noise term,  respectively. Now the objective of HSI restoration is to estimate $\mathcal{X}$ from the observed $\mathcal{Y}$ by exploiting the structures of the clean HSI $\mathcal{X}$ and the noise terms $\mathcal{E}$. 

Following the studies of \cite{he2016total, xie2016hyperspectral}, we divide the noise term 
$\mathcal{E}$ into two sub-terms as is shown in Fig.2, i.e., the Gaussian noise term $\mathcal{N}$ and the sparse noise term $\mathcal{S}$ including stripes, impulse noise, and dead pixels, leading to the following degradation model: 
\begin{equation}
\mathcal{Y}=\mathcal{X}+\mathcal{N}+\mathcal{S}. 
\end{equation}
As such,  the Frobenius norm and the $\ell_1$ norm can be naturally used to model such two noise terms $\mathcal{N}$ and $\mathcal{S}$ respectively. 

It is well known that each spectral signature can be represented by a linear combination of a small number of pure spectral endmembers, which means that the mode-3 matricization $\mathbf{X}_{(3)}$ can be factorized as $\mathbf{X}_{(3)}=\mathbf{M}\mathbf{D}^T$,  where $\mathbf{D}\in\mathbb{R}^{B\times r}$ is the so-called endmember matrix, and $\mathbf{M}\in\mathbb{R}^{wh\times r}$ is regarded as the abundance matrix. As stated in \cite{zhang2014hyperspectral}, the number of endmembers $r$ is relatively small, that is, $r\ll B$ or $r\ll wh$,  which means that only a small fraction of singular values are greater than zero as shown in the singular value curve of $\mathbf{X}_{(3)}$ in Fig. 3. Besides, like other real images, there have certain correlations in both two spatial modes as also shown in Fig. 3, where the singular value curves of $\mathbf{X}_{(1)}$ and $\mathbf{X}_{(2)}$ have obvious decaying trends. Therefore, we can utilize low-rank Tucker decomposition to characterize the aforementioned spatial-and-spectral correlation. 

 TV regularization has been widely used to explore the spatial piecewise smooth structure for tackling HSI restoration task. See, e.g., \cite{he2016total}, \cite{wu2016total}. One can find in Fig. 4 an illustrative example of edge detection for demonstrating the spatial smoothness of one HSI.  Actually, as presented in the third row of Fig. 4, there also exits local smooth structure of a HSI along with its spectral mode. That is, most of the difference value  between adjacency bands in spectral domain nearly equal to 0. It is obvious that the commonly used  band-by-band TV regularizer neglects such spectral smoothness, which motivates us to design a new SSTV regularizer to fully explore the piecewise smooth structure in both spatial and spectral domains. It should be also pointed that the works \cite{chang2015anisotropic, jiang2016hyperspectral} are very different from our work, although they both utilized SSTV regularizers for destriping  and denoising tasks, respectively.

\begin{figure}[!]
\centering
\includegraphics[scale=0.57]{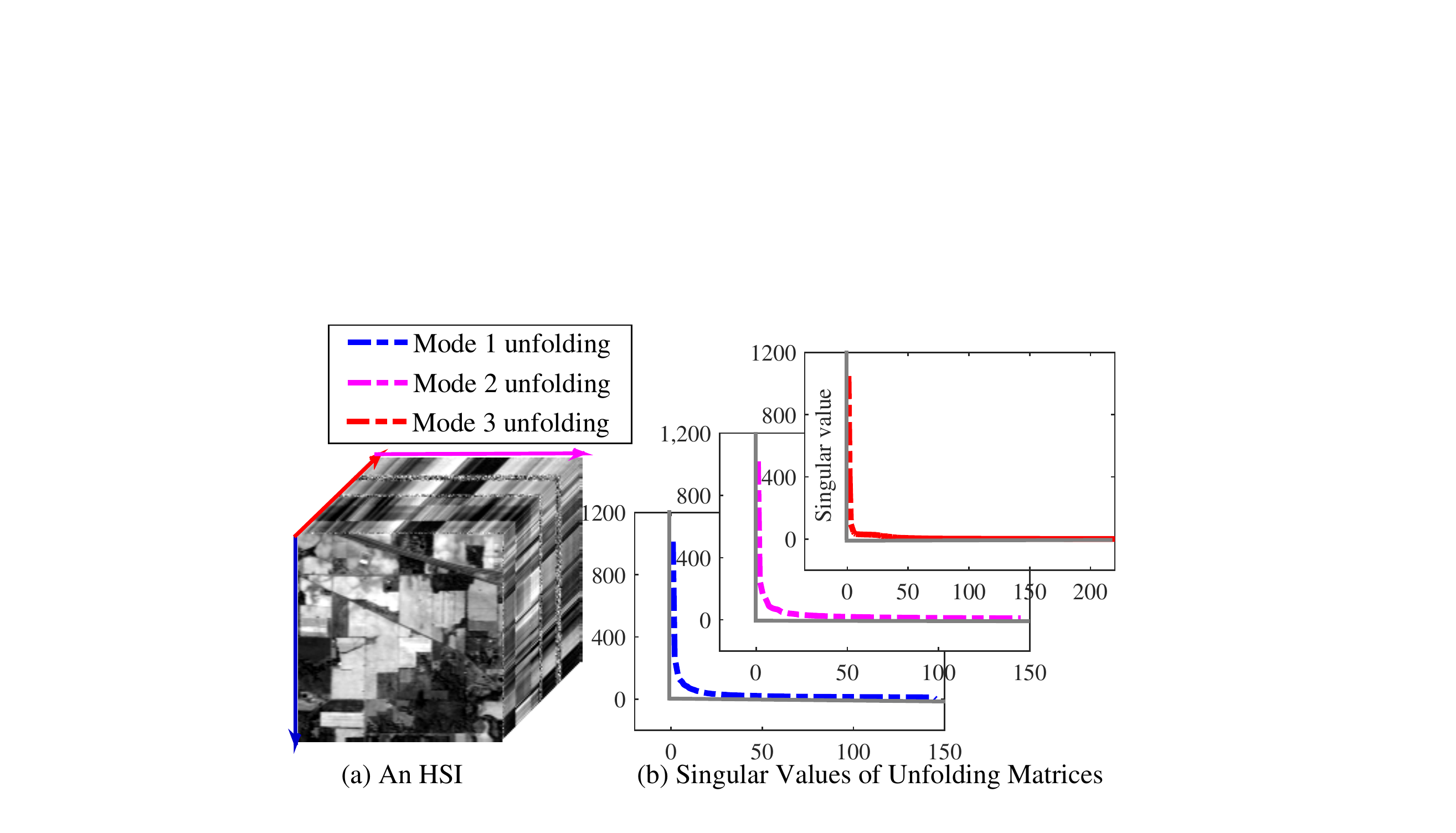}
\caption{The global spatial-and-spectral correlation in HSI cube.}
\label{low-rank prior}
\end{figure}  
\subsection{Low-rank Tensor Decomposition with Anisotropic SSTV}
As previously stated, exploiting the prior knowledge is a key consideration for HSI mixed-noise removal. Based on the discussion of the above Section II.A, by combining the low-rank and TV properties in both spatial and spectral models, we introduce a \textit{TV regularized low-rank tensor decomposition} (LRTDTV for short) model, that is, 
\begin{equation} 
\begin{split}\label{main_model}
&\min_{\mathcal{X},\mathcal{S},\mathcal{N}}\tau \Vert\mathcal{X}\Vert_{\text{SSTV}}+\lambda \Vert\mathcal{S}\Vert_1+\beta \Vert\mathcal{N}\Vert^2_F \\
&~{\rm s.t.}~\mathcal{Y}=\mathcal{X}+\mathcal{S}+\mathcal{N}, \\
& \mathcal{X}=\mathcal{C}\times_1 \mathbf{U}_1\times_2\mathbf{U}_2\times_3\mathbf{U}_3,\mathbf{U}_{i}^{T}\mathbf{U}_{i}=\textbf{I}(i=1,2,3), 
\end{split}
\end{equation}
where $\mathcal{C}\times_1 \mathbf{U}_1\times_2\mathbf{U}_2\times_3\mathbf{U}_3$ is the Tucker decomposition with core tensor $\mathcal{C}$ and factor matrices $\mathbf{U}_is$ of rank $r_is$,  and the SSTV term $\|\mathcal{X}\|_{\text{SSTV}}$ is defined as 
\begin{equation*}
\begin{split}
\|\mathcal{X}\|_{\text{SSTV}}:=&\sum_{i,j,k}w_1 |x_{i,j,k}- x_{i,j,k-1}| + w_2|x_{i,j,k} - x_{i,j-1,k}| \\ 
                                                &+w_3|x_{i,j,k} - x_{i-1,j,k}|, 
\end{split}
\end{equation*}
where $x_{i,j,k}$ is the $(i,j,k)$-th entry of $\mathcal{X}$, and $w_j(j=1, 2, 3)$ is the weight along the $j$-th mode of $\mathcal{X}$ that controls its regularization strength. It is not hard to see that the above problem (\ref{main_model}) is a nonconvex optimization problem due to the nonconvexity of Tucker decomposition. Thus, it would be possible to find a good local solution by using the popular augmented Lagrange multiplier (ALM) method \cite{Lin2009} that we shall show in the next subsection. 

It is worthy noting that the proposed model can fully capture the spatial and spectral information of the HSI,  and thus is expected to have a strong ability of mixed noise removal. Specifically, the Tucker decomposition could make use of the spectral similarity of all the pixels and the certain correlations in both two spatial modes. Once the sparse noise term, including impulse noise, dead lines and stripes, is detected by the $\ell_1$ norm term, the designed SSTV norm term is used to characterize the piecewise smooth structure in both spatial and spectral domains and, as a result, help to remove the Gaussian noise.  The Frobenius norm term is further used to model Gaussian noise and can enhance the performance of the proposed model in some heavy Gaussian noise situations.



\begin{figure}[!]
\centering
\includegraphics[scale=0.42]{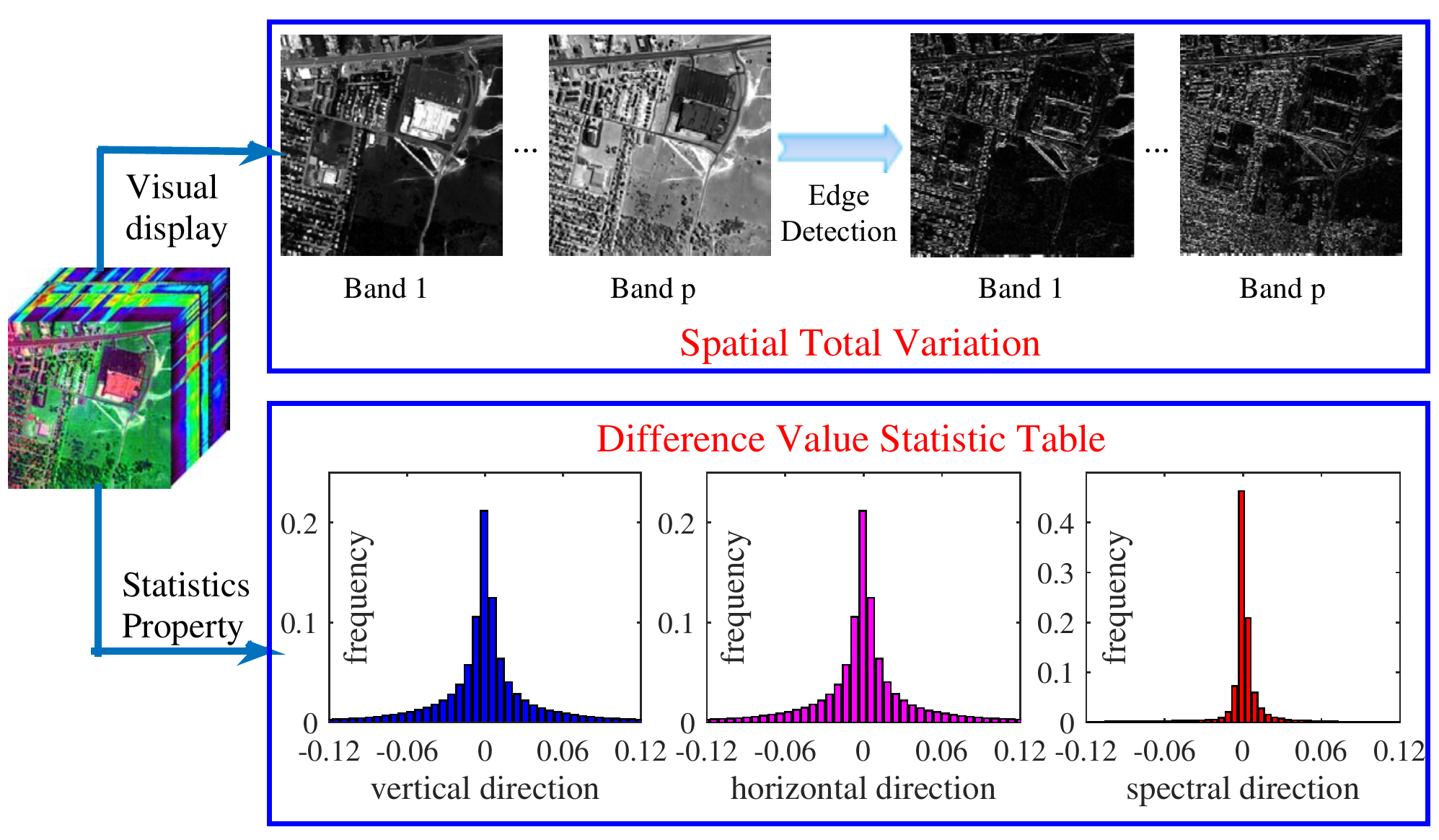}
\caption{The property of total variation of HSI in both spatial and spectral modes.}
\label{SSTV}
\end{figure} 

\subsection{Optimization Procedure}
By introducing some  auxiliary variables , we rewrite  (\ref{main_model}) as  the following equivalent  minimization problem:
\begin{equation}\label{alm1}
\begin{split}
&\min_{\mathcal{C}, \mathbf{U}_{i}, \mathcal{X}, \mathcal{F}, \mathcal{S}, \mathcal{N}} \tau\Vert\mathcal{F}\Vert_1+\lambda \Vert\mathcal{S}\Vert_1+\beta \Vert\mathcal{N}\Vert^2_F\\
&~{\rm s.t.}~\mathcal{Y}=\mathcal{X}+\mathcal{S}+\mathcal{N},\mathcal{X}=\mathcal{Z},{D_w}(\mathcal{Z})=\mathcal{F},\\
&\qquad \mathcal{X}=\mathcal{C}\times_1 \mathbf{U}_1\times_2\mathbf{U}_2\times_3\mathbf{U}_3,\mathbf{U}_{i}^{T}\mathbf{U}_{i}=\mathbf{I}, 
\end{split}
\end{equation}
where ${D_w}(\cdot)=\left[{w}_1\times{D_h}(\cdot);{w}_2 \times{D_v}(\cdot);{w}_3\times {D_t}(\cdot)\right]$ is the so-called weighted three-dimensional difference operator and ${D_h}$, $D_v$, $D_t$ are the first-order difference operators respect to three different directions of HSI cube. Based on the ALM methodology, the above problem (\ref{alm1}) can be transformed into minimizing  the following augmented Lagrangian function: 
\begin{equation}\label{alm2}
\begin{split}
&{L}(\mathcal{X}, \mathcal{S}, \mathcal{N}, \mathcal{Z}, \mathcal{F}, \Gamma_1, \Gamma_2, \Gamma_3)= 
\tau \Vert\mathcal{F}\Vert_1+\lambda \Vert\mathcal{S}\Vert_1+\beta \Vert\mathcal{N}\Vert^2_F\\
&\qquad\left\langle \Gamma_1, \mathcal{Y}-\mathcal{X}-\mathcal{S}-\mathcal{N}\right\rangle+\left\langle \Gamma_2, \mathcal{X}-\mathcal{Z}\right\rangle \\
&\qquad +\left\langle \Gamma_3, {D_w}(\mathcal{Z})-\mathcal{F}\right\rangle +\frac{\mu}{2}\left(\Vert\mathcal{Y}-\mathcal{X}-\mathcal{S}-\mathcal{N}\Vert^2_F\right. \\
&\qquad\left. +\Vert\mathcal{X}-\mathcal{Z}\Vert^2_F+\Vert{D_w}(\mathcal{Z})-\mathcal{F}\Vert^2_F\right),
\end{split}
\end{equation}
under the constraints $\mathcal{X}=\mathcal{C}\times_1 \mathbf{U}_1\times_2\mathbf{U}_2\times_3\mathbf{U}_3$,  and $\mathbf{U}_{i}^{T}\mathbf{U}_{i}=\textbf{I}$, where $\mu$ is the penalty parameter, and $\Gamma_i(i=1,2,3)$ are the Lagrange multipliers.  Therefore, we can alternative optimize the augmented Lagrangian function (\ref{alm2}) over one variable, while fixing the others. Specifically, in the $(k+1)$th iteration, variables involved in the model (\ref{main_model}) can be updated as follows:

1) \textbf{Update} $\mathbf{\mathcal{C}},\mathbf{U}_{i},\mathbf{\mathcal{X}}$. Extracting all terms containing $\mathcal{X}$ from the augmented Lagrangian function (\ref{alm2}), we need to solve:
\begin{equation*}
\begin{split}
&\mathop{\min}_{\begin{subarray}{c}\mathbf{U}_{i}^{T}\mathbf{U}_{i}=\textbf{I}, \\
                    \mathcal{X}=\mathcal{C}\times_1 \mathbf{U}_1\times_2\mathbf{U}_2\mathbf\times \mathbf{U}_3
         \end{subarray}}\langle \Gamma_1^{(k)}, \mathcal{Y}-\mathcal{X}-\mathcal{S}^{(k)}-\mathcal{N}^{(k)} \rangle + \langle \Gamma_2^{(k)},\\
 &\mathcal{X}-\mathcal{Z}^{(k)}\rangle+\frac{\mu}{2}\Big(\Vert\mathcal{Y}-\mathcal{X}-\mathcal{S}^{(k)}-\mathcal{N}^{(k)}\Vert^2_F+\Vert\mathcal{X}-\mathcal{Z}^{(k)}\Vert^2_F\Big), 
 \end{split}
\end{equation*}
which can be easily transformed into the following equivalent problem:
\begin{equation*}
\begin{split}
&\mathop{\min}_{\mathbf{U}_{i}^{T}\mathbf{U}_{i}=\textbf{I}}\mu\Big\Vert\mathcal{C}\times_1 \mathbf{U}_1\times_2\mathbf{U}_2\times_3\mathbf{U}_3-\frac{1}{2} \Big(\mathcal{Y}-\mathcal{S}^{(k)}-\mathcal{N}^{(k)} \\
&\qquad +\mathcal{Z}^{(k)}+(\Gamma_1^{(k)}-\Gamma_2^{(k)})/\mu\Big)\Big\Vert^2_F.
\end{split}
\end{equation*}
By using the classic HOOI algorithm~\cite{K2009Tensor}, we can easily get the  $\mathcal{C}^{(k+1)}$, and $\textbf{U}_i^{(k+1)}(i=1,2,3)$. Then $\mathcal{X}$ can be updated as follows:
\begin{equation}
\mathcal{X}^{(k+1)}=\mathcal{C}^{(k+1)}\times_1 \mathbf{U}_1^{(k+1)}\times_2\mathbf{U}_2^{(k+1)}\times_3\mathbf{U}_3^{(k+1)}. 
\end{equation}
2) \textbf{Update} ${\mathcal{Z}}$. Extracting all terms containing $\mathcal{Z}$ from the augmented Lagrangian function (\ref{alm2}), we can deduced:
\begin{equation*}
\begin{split}
&\quad \mathcal{Z}^{(k+1)}=\mathop{\argmin}_{\mathcal{Z}} \langle \Gamma_2^{(k)},\mathcal{X}^{(k+1)}-\mathcal{Z} \rangle +\langle \Gamma_3^{(k)}, {D_w}(\mathcal{Z})-\\
&\quad\quad \mathcal{F}^{(k)} \rangle+ \frac{\mu}{2}\left(\Vert\mathcal{X}^{(k+1)}-\mathcal{Z}\Vert^2_F+\Vert{D_w}(\mathcal{Z})-\mathcal{F}^{(k)}\Vert^2_F\right). \\
\end{split}
\end{equation*}
Thus, optimizing this problem can be treated as solving the following linear system:
$$(\mu\mathbf{I}+\mu{D^*_w}{D_w})\mathcal{Z} = \mu \mathcal{X}^{(k+1)}+\mu{D^*_w}(\mathcal{F}^{(k)})+\Gamma_2^{(k)}-{D^*_w}(\Gamma_{3}^{(k)}), $$
where ${D^*_w}$ indicates the adjoint operator of $D_w$. Thanks to the block-circulant structure of the matrix corresponding to the operator ${D^*_w}{D_w}$, it can be diagonalized by the 3{D} FFT matrix. Therefore, we have  
\begin{equation}
\left\{
\begin{split}
&\text{H}_z=\mu\mathcal{X}^{(k+1)}+\mu{D^*_w}(\mathcal{F}^{(k)})+\Gamma_2^{(k)}-{D^*_w}(\Gamma_{3}^{(k)}), \\
&\text{T}_z=\text{w}_1^2|\text{fftn}({D}_h)|^2+\text{w}_2^2|\text{fftn}({D}_v)|^2+\text{w}_3^2|\text{fftn}({D}_t)|^2, \\
&\mathcal{Z}^{(k+1)}=\text{ifftn}\left(\frac{\text{fftn}(\text{H}_z)}{\mu\mathbf{1}+\mu\text{T}_z}\right), 
\end{split}
\right.
\end{equation}
  where fftn and ifftn indicate fast $3\mathbf{D}$ Fourier transform and its inverse transform respectively, $|\cdot|^2$ is the elements-wise square, and the division is also performed element-wisely.
  
3) \textbf{Update} ${\mathcal{F}}$. Extracting all terms containing $\mathcal{F}$ from (\ref{alm2}), we can get 
\begin{equation*}
\begin{split}
&\mathcal{F}^{(k+1)}=\mathop{\argmin}_{\mathcal{F}} \tau\Vert F\Vert_1+\left\langle \Gamma_3^{(k)}, {D_w}(\mathcal{Z}^{(k+1)})-\mathcal{F} \right\rangle \\
& +\frac{\mu}{2}||{D_w}(\mathcal{Z}^{(k+1)})-\mathcal{F}||^2_F\\
&=\mathop{\argmin}_{\mathcal{F}} \tau\Vert\mathcal{F}\Vert_1+\frac{\mu}{2}\Vert\mathcal{F}-({D_w}(\mathcal{Z}^{(k+1)})+\frac{M_3^{(k)}}{\mu})\Vert^2_F.
\end{split}
\end{equation*}
By introducing the so-called \textit{soft-thresholding operator}:
\begin{equation*}
\mathcal{R}_\Delta(\mathbf{x})=\left\{
\begin{split}
{x}-\Delta, &   & if \quad\mathbf{x} > \Delta  \\
{x}+\Delta, &   & if \quad\mathbf{x} < \Delta \\
0,                 &   & \qquad otherwise
\end{split}
\right.
\end{equation*}
where $ x\in\mathbb{R}$ and $ \Delta >0 $, then we can update $\mathcal{F}^{k+1}$ as
\begin{equation}
\mathcal{F}^{k+1}=\mathcal{R}_{\frac{\tau}{\mu}}\Big({D_w}(\mathcal{Z}^{(k+1)})+\frac{\Gamma_3^{(k)}}{\mu} \Big). 
\end{equation}
4) \textbf{Update} ${\mathcal{S}}$. Similarly,  we should consider 
\begin{align*}
&\mathcal{S}^{(k+1)}=\mathop{\argmin}_{\mathcal{S}} \lambda\Vert\mathcal{S}\Vert_1+\langle \Gamma_1^{(k)}, \mathcal{Y}-\mathcal{X}^{(k+1)}-\mathcal{S}-\mathcal{N}^{(k)} \rangle\\
&+\frac{\mu}{2}\Vert\mathcal{Y}-\mathcal{X}^{(k+1)}-\mathcal{S}-\mathcal{N}^{(k)}\Vert_F^2\\
&=\mathop{\argmin}_{\mathcal{S}} \lambda\Vert\mathcal{S}\Vert_1+\frac{\mu}{2}\Big\Vert\mathcal{S}-(\mathcal{Y}-\mathcal{X}^{(k+1)}-\mathcal{N}^{(k)}+\frac{\Gamma_1^{(k)}}{\mu})\Big\Vert_F^2.
\end{align*}
By using the soft-thresholding operator introduced before, then the solution of above problem can be expressed as 
\begin{equation}
\mathcal{S}^{k+1}=\mathcal{R}_{\frac{\lambda}{\mu}}\Big(\mathcal{Y}-\mathcal{X}^{(k+1)}-\mathcal{N}^{(k)}+\frac{M_1^{(k)}}{\mu}\Big).
\end{equation}
5) \textbf{Update} ${\mathcal{N}}$. Extracting all items related to $\mathcal{N}$ from (\ref{alm2}), it can be easily deduced:
\begin{equation*}
\begin{split}
&\mathcal{N}^{(k+1)}=
\mathop{\argmin}_{\mathcal{N}} \beta\Vert\mathcal{N}\Vert^2_F+\langle \Gamma_1^{(k)}, \mathcal{Y}-\mathcal{X}^{(k+1)}-\mathcal{S}^{(k+1)}-\mathcal{N}\rangle\\
&+\frac{\mu}{2}\Vert\mathcal{Y}-\mathcal{X}^{(k+1)}-\mathcal{S}^{(k+1)}-\mathcal{N}\Vert^2_F\\
&=\mathop{\argmin}_{\mathcal{N}} (\beta+\frac{\mu}{2})\Big\Vert\mathcal{N}-\frac{\mu\left(\mathcal{Y}-\mathcal{X}^{(k+1)}-\mathcal{S}^{(k+1)}\right)+\Gamma_{1}^{(k)}}{\mu+2\beta}\Big\Vert^2_F.
\end{split}
\end{equation*}
Thus through a simple calculation,  one can get its solution as follows:
\begin{equation}
\mathcal{N}^{k+1}=\frac{\mu\left(\mathcal{Y}-\mathcal{X}^{(k+1)}-\mathcal{S}^{(k+1)}\right)+M_{1}^{(k)}}{\mu+2\beta}
\end{equation}
6) \textbf{Updating Multipliers}.  Based on the ALM algorithm, the multipliers are updated by the following equations: 
\begin{equation}
\left\{
\begin{split}
&\Gamma_1^{(k+1)}=\Gamma_1^{(k)}+\mu\Big(\mathcal{Y}-\mathcal{X}^{(k+1)}-\mathcal{S}^{(k+1)}-\mathcal{N}^{(k+1)}\Big)\\
&\Gamma_2^{(k+1)}=\Gamma_2^{(k)}+\mu\Big(\mathcal{X}^{(k+1)}-\mathcal{Z}^{(k+1)}\Big)\\
&\Gamma_3^{(k+1)}=\Gamma_3^{(k)}+\mu\Big({D_w}(\mathcal{Z}^{(k+1)})-\mathcal{F}^{(k+1)}\Big).
\end{split}
\right.
\end{equation}

Summarizing the aforementioned discussion, we now arrive at obtaining an ALM method to solve the proposed LRTDTV model (\ref{main_model}), as presented in Algorithm 1.
\begin{algorithm}[htp]
 \caption{LRTDTV solver}
 \label{LRDTV_algorihm}
   \begin{algorithmic}[1]
   \REQUIRE The noisy HSI $\mathcal{Y}$, desired rank $[r_{1},r_{2},r_{3}]$, stopping criteria $\epsilon$, and the regularization parameters $\tau $, $\lambda$ and $\beta$, and the weights $w_i$s. 
   \ENSURE The restored HSI $\mathcal{X}$
   \STATE Initialize $\mathcal{X}=\mathcal{Z}=\mathcal{S}=\mathcal{N}=0$, $\Gamma_{1}=\Gamma_{2}=\Gamma_{3}=0$, $\mu_{max}=10^6$, $\rho=1.5$, and ${k}=0$
   \STATE Repeat until convergence \\
            Update $\mathbf{\mathcal{X}},\mathbf{\mathcal{Z}},F,\mathbf{\mathcal{S}},\mathbf{\mathcal{N}},\Gamma_{1} ,\Gamma_{2}, \Gamma_{3}$  via (6-11) \\ 
     Update the parameter $\mu :=\min ({\rho\mu,\mu_{max}})$ \\ 
    Check the convergence condition\\
   $\qquad\qquad \frac{||\mathcal{X}^{(k)}-\mathcal{X}^{(k+1)}||^2_F}{||\mathcal{Y}||^2_F}\leq \epsilon$
   \end{algorithmic}
 \end{algorithm}

In the LRTDTV solver, the inputs are the noisy image $\mathcal{Y}\in\mathbb{R}^{M\times N\times p}$, desired rank $[r_{1},r_{2},r_{3}]$ for Tucker decomposition, the stopping criteria $\epsilon$, and the regularized parameters $\tau $, $\lambda$, and $\beta$. Considering the fact that these three parameters have certain proportional relationship, we simply set $\tau=1$ and then tune $\lambda$ and $\beta$. More details and discussions would be presented in Section IV.C. For another important parameter $\mu$,  we first initialize it as $\mu=10^{-2}$ and then update it as $\mu=\min (\rho\mu,\mu_{max})$ in each iteration. This strategy of determining the variable $\mu$ has been widely used in the ALM-based methods \cite{Lin2009, bertsekas2014constrained}, which can facilitate the convergence of the algorithm.

\subsection{The LRTDTV Approximate Model}
In the LRTDTV model (\ref{main_model}), we separate the mixture noise into two parts namely the sparse noise and the Gaussian noise. Owing to the TV regularization has certain effect to remove the Gaussian noise, when the the variance of Gaussian noise is not heavy, we can simplify (\ref{main_model}) as follows:
\begin{equation}\label{app_model}
\begin{split}
&\min_{\mathcal{X},\mathcal{S}}\tau\Vert\mathcal{X}\Vert_{\text{SSTV}}+\lambda \Vert\mathcal{S}\Vert_1\\
&~{\rm s.t.}~\mathcal{Y}=\mathcal{X}+\mathcal{S} \\
& \mathcal{X}=\mathcal{C}\times_1 \mathbf{U}_1\times_2\mathbf{U}_2\times_3\mathbf{U}_3,\mathbf{U}_{i}^{T}\mathbf{U}_{i}=\mathbf{I}(i=1,2,3). 
\end{split}
\end{equation}
Similarly, we can derive an ALM-based optimization procedure. For simplicity, we directly give the closed-form solution of each subproblem in the following.

\textbf{Solving} ${\mathcal{X}}$ subproblem:

We need to consider the following optimization problem:
\begin{equation*}
\begin{split}
&\mathop{\min}_{\mathbf{U}_{i}^{T}\mathbf{U}_{i}=\mathbf{I}}\mu\Big\Vert\mathcal{C}\times_1 \mathbf{U}_1\times_2\mathbf{U}_2\times_3\mathbf{U}_3-\frac{1}{2} \Big(\mathcal{Y}-\mathcal{S}^{(k)}+\mathcal{Z}^{(k)} \\
&\qquad\qquad +(\Gamma_1^{(k)}-\Gamma_2^{(k)})/\mu\Big)\Big\Vert^2_F.
\end{split}
\end{equation*}
By using HOOI algorithm, we can get the estimation of the core tensor $\mathcal{C}^{(k)}$, and $\mathbf{U}_i^{(k)}(i=1,2,3)$. Then $\mathcal{X}^{(k+1)}$ can be updated as follows:
\begin{equation}
\mathcal{X}^{(k+1)}=\mathcal{C}^{(k+1)}\times_1 \text{U}_1^{(k+1)}\times_2\text{U}_2^{(k+1)}\times_3\text{U}_3^{(k+1)}.
\end{equation}

\textbf{Solving} ${\mathcal{Z}}$ subproblem:
\begin{equation}
\left\{
\begin{split}
&\text{H}_z=\mu\mathcal{X}^{(k+1)}+\mu{D^*_w}(\mathcal{F}^{(k)})+\Gamma_2^{(k)}-{D^*_w}(\Gamma_{3}^{(k)}) \\
&\text{T}_z=\text{w}_1^2|\text{fftn}({D}_h)|^2+\text{w}_2^2|\text{fftn}({D}_v)|^2+\text{w}_3^2|\text{fftn}({D}_t)|^2 \\
&\mathcal{Z}^{(k+1)}=\text{ifftn}\left(\frac{\text{fftn}(\text{H}_z)}{\mu\mathbf{1}+\mu\text{T}_z}\right).
\end{split}
\right.
\end{equation}

\textbf{Solving} ${\mathcal{F}}$ subproblem:
\begin{equation}
\mathcal{F}^{k+1}=\mathcal{R}_{\frac{\tau}{\mu}}\Big({D_w}(\mathcal{Z}^{(k+1)})+\frac{\Gamma_3^{(k)}}{\mu} \Big).
\end{equation}

\textbf{Solving} ${\mathcal{S}}$ subproblem:
\begin{equation}
\qquad\mathcal{S}^{k+1}=\mathcal{R}_{\frac{\lambda}{\mu}}\Big(\mathcal{Y}-\mathcal{X}^{(k+1)}+\frac{\Gamma_1^{(k)}}{\mu}\Big).
\end{equation}

The multipliers are updated by the following equations:
\begin{equation}
\left\{
\begin{aligned}
&\Gamma_1^{(k+1)}=\Gamma_1^{(k)}+\mu\Big(\mathcal{Y}-\mathcal{X}^{(k+1)}-\mathcal{S}^{(k+1)} \Big)\\
&\Gamma_2^{(k+1)}=\Gamma_2^{(k)}+\mu\Big(\mathcal{X}^{(k+1)}-\mathcal{Z}^{(k+1)}\Big)\\
&\Gamma_3^{(k+1)}=\Gamma_3^{(k)}+\mu\Big({D_w}(\mathcal{Z}^{(k+1)})-\mathcal{F}^{(k+1)}\Big).
\end{aligned}
\right.
\end{equation}

Similarly, we could get an efficient ALM method to solve the LRTDTV approximate model (\ref{app_model}), that is, sequentially update (13-16) until certain convergence condition is satisfied.  Note that all the parameters used for updating (13-16) can be set as the same values that used in Algorithm 1. 

 
\section{EXPERIMENTAL RESULTS AND DISCUSSION}
Both simulated and real image data experiments were carried out to demonstrate the effectiveness of the proposed LRTDTV method for HSI restoration. To thoroughly evaluate the performance of LRTDTV, we implemented seven different popular HSI restoration methods for comparison, i.e., the nuclear norm minimazition (NNM) based LRMA \cite{Candes2011rpca}, the weighted nuclear norm minimazition (WNNM) based LRMA \cite{gu2014weighted}, the GoDec based low-rank matrix recovery (LRMR) \cite{zhang2014hyperspectral}, the  weighted schatten norm minimazition (WSNM) based LRMA\cite{xie2016hyperspectral}, the TV-regularized low-rank matrix factorization (LRTV)\cite{he2016total}, the block matching 4-D filtering (BM4D) \cite{Mag2012Nonlocal}, and the decomposable nonlocal tensor dictionary learning (TDL) \cite{peng2014decomposable}. These methods represent the state-of-the-art HSI restoration methods, especially WSNM and LRTV,  and their implementation codes can be directly obtained from the authors' websites. 

 In all the following experiments, the parameters in those compared methods were manually adjusted according to their default strategies. While for our LRTDTV solver,  we would like to present a detailed discussion of parameter selection in Section IV.C. In addition, to facilitate the numerical calculation and visualization, all the bands of the HSI are normalized into $[0, 1]$ and then will be stretched to original level after restoration. 
 
\subsection{Simulated Data Experiments}
The synthetic data were firstly considered by \cite{zhang2014hyperspectral}, which were generated using the ground truth of the Indian Pines dataset \cite{Indian}. The size of the synthetic HSI was $145\times 145\times 224$, and the reflectance values of all the voxels in the HSI were linearly mapped to $[0, 1]$. 

To simulate noisy HSI data, we added several types of noise to the original synthetic HSI data under six different cases to test the performance of all the compared restoration methods, both in visual quality and quantitative perspective. We list the details in the following:


{Case 1)}: For different bands, the noise intensity was equal. And the same distribution of zero-mean Gaussian noise was added to each band. The variances of the Gaussian noise was 0.1.

{Case 2)}: In this case, the Gaussian noise was added to each band just as in Case 1). In addition, some deadlines were added from  band 91 to band 130, with the number of stripes was randomly selected from 3 to 10,  and the width of the stripes was randomly generated from 1 to 3.

{Case 3)}: In this case, the noise intensity was equal for different bands. And the same distribution of zero-mean Gaussian noise and the same percentage of impulse noise were added to each band. The variance of the Gaussian noise was 0.075, and the percentage of the impulse noise was 0.15.

{Case 4)}: In this case, the Gaussian and impulse noise were added just like in Case 3). Besides, the deadlines were added from band 91 to band 130, with the number of stripes was randomly selected from 3 to 10, and the width of the deadlines was randomly generated from 1 to 3.

{Case 5)}: In this case, the noise intensity was different for different bands. That is, different variance zero-mean Gaussian noise was added to each band, with the variance value being randomly selected from 0 to 0.2, and different percentages of impulse noise were added, which were randomly selected from 0 to 0.2. In addition, the deadlines noise were added to some band just as in Case 4).

{Case 6)}: In this case, the Gaussian noise, impulse noise, and deadlines were added just as in Case 5). In addition,  some stripes were added from band 161 to band 190, with the number of stripes being randomly selected from 20 to 40.

\subsubsection{{Visual quality comparison}}

In terms of visual quality, two representative bands of restored HSIs in two typical cases obtained by different methods are shown in Fig. 5 and Fig. 6, respectively. More precisely, Fig. 5 shows the restoration results of band 36, which is corrupted by both the Gaussian noise  and impulse noise in Case 3). And Fig. 6 shows the restoration results of band 116, which is severely corrupted by a mixture of Gaussian noise, impulse noise and deadline in Case 5). 
  \begin{figure*}[!]
\centering
\includegraphics[scale=0.7]{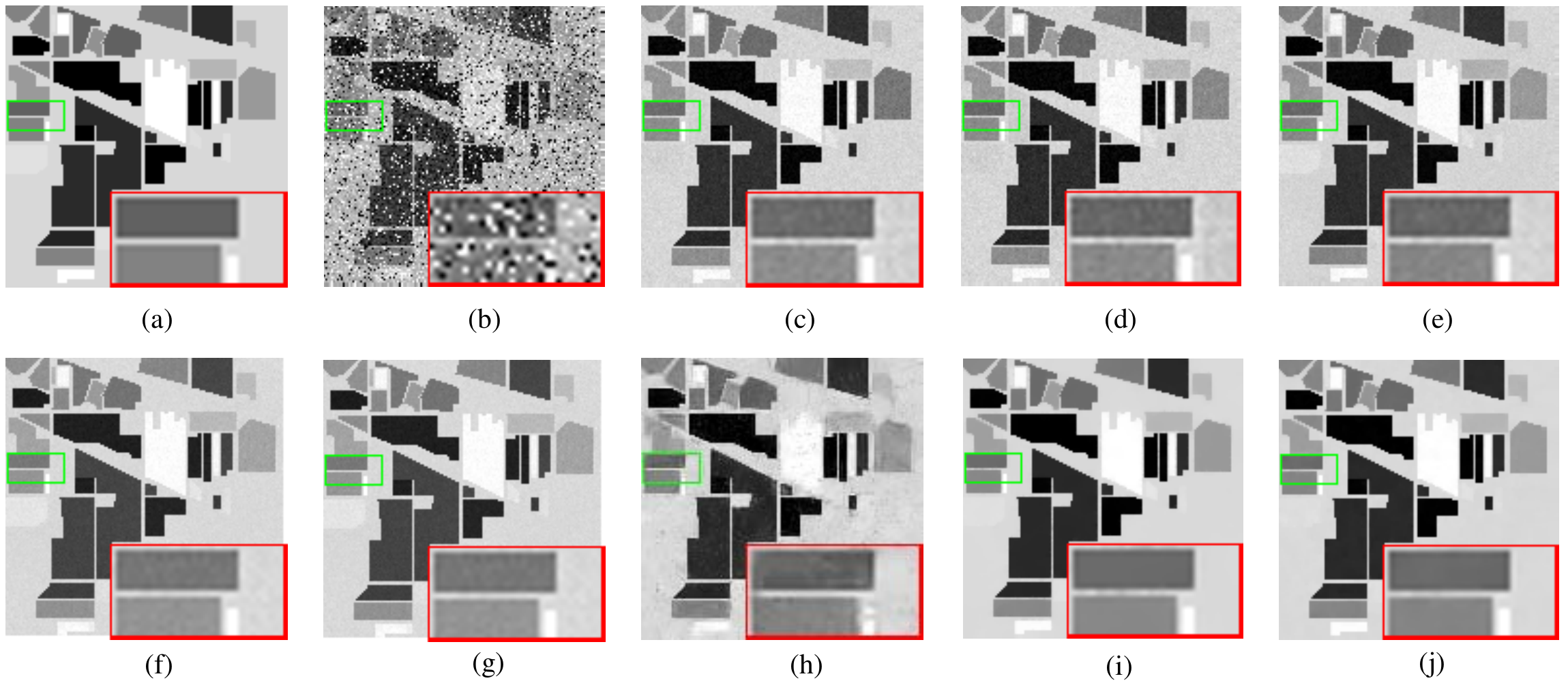}
\caption{Denoised results by all the compared methods: (a) the original band $36$, (b) the simulated noise band of Case 3), (c) NNM, (d) WNNM, (e) LRMR, (f) BM4D, (g) TDL, (h) WSNM, (i) LRTV, (j) LRTDTV. }
\label{case 3)}
\end{figure*} 

It is evident that, for both cases, the original clean HSIs have undergone great changes as shown in Fig. 5(b) and Fig. 6(b) respectively. Aiming at better visual comparison,  we marked a same subregion of each subfigure in Figs. 5 and 6 by a green box and then enlarged it in a red box. Several observations can be easily made from both Figs. 5 and 6. Firstly, all the compared methods, to some extent, could remove such mixed-noise. Secondly, the proposed LRTDTV method performs best among all the compared methods, effectively removing the noise while preserving the essential structures of original HSIs. For instance,  compared to red dashed boxes in Figs. 6(c-i), the red dashed box in Fig. 6(j) produced by our LRTDTV method preserves the edge clearer and sharper. Thirdly, the two TV-regularized methods, i.e., LRTV and LRTDTV, perform much better than other competing methods, and our proposed LRTDTV out-performs LRTV to a certain extent. This demonstrates the power of both TV regularization and directly low rank tensor modeling techniques for HSI restoration. 

\begin{figure*}[!]
\centering
\includegraphics[scale=0.7]{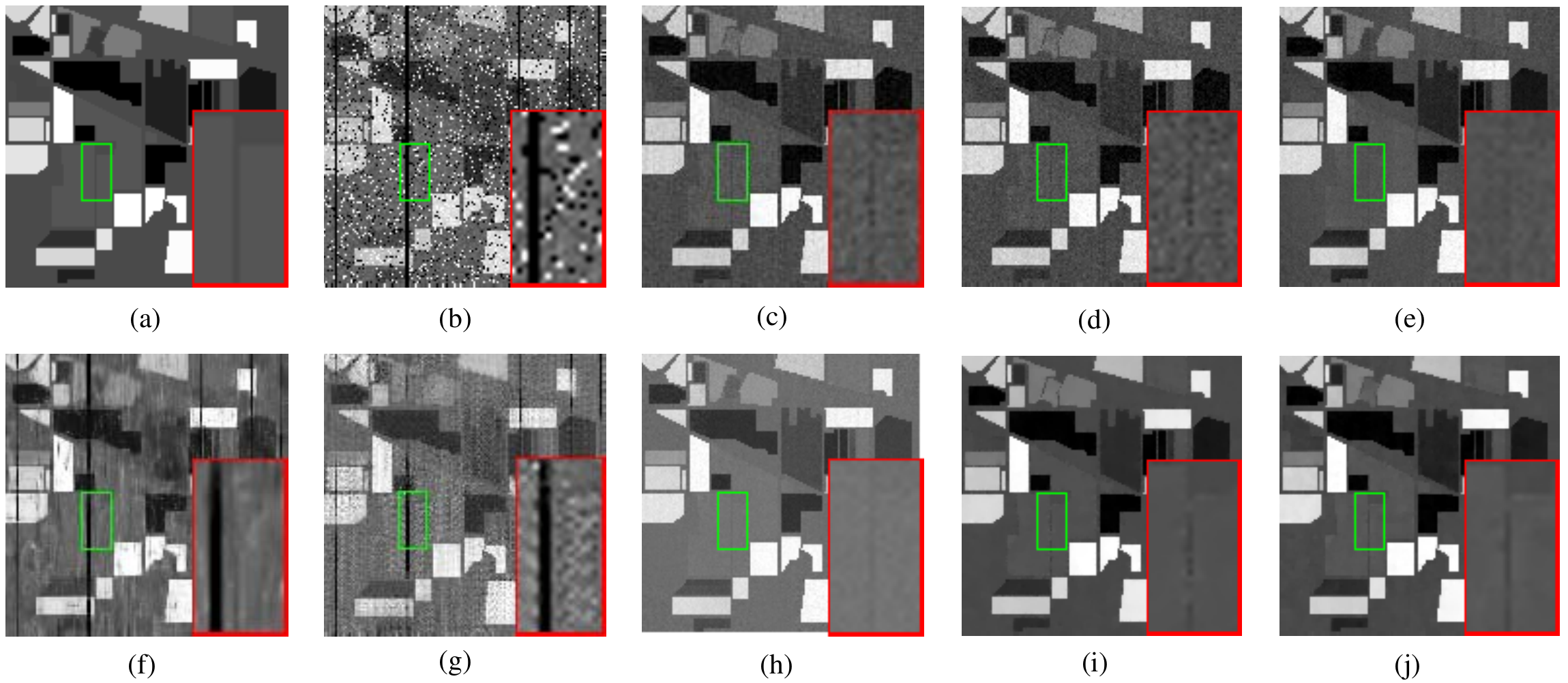}
\caption{Denoised results by all the compared methods: (a) the original band $116$, (b) the simulated noise band of Case 5), (c) NNM, (d) WNNM, (e) LRMR, (f) BM4D, (g) TDL, (h) WSNM, (i) LRTV,   (j) LRTDTV. }
\label{case 5)}
\end{figure*}   
    

\subsubsection{{Quantitative comparison}}
To  better understand the overall performance of different methods, the mean peak signal-to-noise ratio (MPSNR) and mean structural similarity index (MSSIM) and Erreur Relative Globale Adimensionnelle de Synthese (ERGAS) are adopted to objectively evaluate the restoration quality:
\begin{align*}
\text{MPSNR} = \frac{1}{B}\sum_{i=1}^B \text{psnr}_i, \\
\text{MSSIM} = \frac{1}{B}\sum_{i=1}^B \text{ssim}_i, \\
\text{ERGAS} = \sqrt{\frac{1}{B}\sum_{i=1}^B\frac{mse(\text{ref}_i,\text{res}_i)}{Mean_2(\text{ref}_i)}}, 
\end{align*}
where  $\text{psnr}_i$ and $\text{ssim}_i$ denote the PSNR and SSIM values for the $i$th band, $\text{ref}_i$ and $\text{res}_i$ denote the $i$th band of the reference image  the restoration image, respectively.

Table II documents the restoration results by all the compared methods in terms of aforementioned three indices. Because the Gaussian noise is not the dominant noise in Case 2)-6), so we adopt  the approximation model (15) in such cases except Case 1) where the Gaussian noise dominates other noises. It is clear that the proposed LRTDTV enjoys a superior performance over the other popular approaches. We also show in Fig 7.  that the PSNR and SSIM values of each band in all the simulated data cases. It is easy to see that, the TV-regularized methods including LRTV and the proposed LRTDTV can get much higher SSIM and PSNR values than other ones for almost every band. In addition, except that the SSIM values of LRTV are slightly higher than those of LRTDTV in Case 1), LRTDTV achieves the best performance among all the methods in terms of both SSIM and PSNR .


\begin{table*}
\caption{QUANTITATIVE EVALUATION OF DIFFERENT METHODS ON THE SIMULATED DATA IN DIFFERENT CASES}
\centering
\begin{tabular}{c c c c c c c c c c c } 
\Xhline{4\arrayrulewidth}
\multicolumn{1}{c}{$\text{Noise case}$} & \multicolumn{1}{c}{$\text{Evaluation index}$} &\multicolumn{1}{c}{$\text{Noise}$} &\multicolumn{1}{c}{$\text{NNM}$} &\multicolumn{1}{c}{$\text{WNNM}$} &\multicolumn{1}{c}{$\text{LRMR}$}& \multicolumn{1}{c}{$\text{BM4D}$}&\multicolumn{1}{c}{$\text{TDL}$}& \multicolumn{1}{c}{$\text{WSNM}$} & \multicolumn{1}{c}{$\text{LRTV}$} & \multicolumn{1}{c}{$\text{LRTDTV}$} \\
\Xhline{3\arrayrulewidth}
\multirow{3}{*}{Case 1)} & MPSNR(dB) &19.99 &30.97 &32.58 &36.20 &38.44 &38.05 &37.32 &38.68 & \textbf{40.76} \\
& MSSIM &0.3672 &0.8727 &0.8420 &0.9311 &0.9763 &0.9674 &0.9453 &\textbf{0.9853} &0.9804\\
& ERGAS &233.99 &69.02 &57.65 &36.85 &29.04 &30.72 &32.37 &28.47  &\textbf{23.02}\\
\Xhline{2\arrayrulewidth}
\multirow{3}{*}{Case 2)} & MPSNR(dB) &19.34 &30.81 &32.46 &35.67 &35.10 &33.22 &36.12 &38.04  & \textbf{40.54}\\
& MSSIM &0.3592 &0.8715 &0.8415 &0.9291 &0.9391 &0.8743 &0.9402 &0.9818 &\textbf{0.9895}\\
& ERGAS &257.88 &70.06 &58.39 &39.84 &110.40 &112.35 &44.89 &49.18  & \textbf{23.44}\\
\Xhline{2\arrayrulewidth}
\multirow{3}{*}{Case 3)} & MPSNR(dB) &13.07 &31.61 &32.36 &36.40 &28.59 &27.50 &38.14 &39.54 & \textbf{41.08}\\
& MSSIM &0.1778 &0.8889 &0.8786 &0.9345 &0.8401 &0.9076 &0.9551 &0.9866 &\textbf{0.9910}\\
& ERGAS &520.53 &64.36 &59.71 &36.04 &90.29 &102.08 &29.52 &35.22 & \textbf{21.98}\\
\Xhline{2\arrayrulewidth}
\multirow{3}{*}{Case 4)} & MPSNR(dB) &12.92 &31.45 &32.29 &35.76 &27.02 &26.54 &36.63 &38.75 & \textbf{40.72}\\
& MSSIM &0.1748 &0.8878 &0.8777 &0.9316 &0.8073 &0.8365 &0.9485 &0.9826 & \textbf{0.9906}\\
& ERGAS &529.82 &65.45 &60.10 &39.99 &128.11 &120.54 &46.32 &55.44  & \textbf{22.90}\\
\Xhline{2\arrayrulewidth}
\multirow{3}{*}{Case 5)} & MPSNR(dB) &13.80 &29.62 &31.33 &33.72 &27.95 &24.34 &35.02 &36.54  & \textbf{38.83}\\
& MSSIM &0.2038 &0.8633 &0.8445 &0.8951 &0.8201 &0.5931 &0.9285 &0.9742  &\textbf{0.9859}\\
& ERGAS &500.68 &81.16 &66.39 &50.16 &121.95 &158.75 &51.33 &72.52  & \textbf{28.66}\\
\Xhline{2\arrayrulewidth}
\multirow{3}{*}{Case 6)} & MPSNR(dB) &13.73 &29.54 &29.97 &33.42 &27.53 &23.34 &33.88 &36.35  & \textbf{38.63}\\
& MSSIM &0.2022 &0.8612 &0.8431 &0.8918 &0.8060 &0.5583 &0.9261 &0.9736  &\textbf{0.9852}\\
& ERGAS &504.37 &82.18 &82.48 &52.62 &127.28 &174.28 &53.29 &72.05  & \textbf{29.82}\\
\Xhline{4\arrayrulewidth}
\end{tabular}
\label{table:t1}
\end{table*}

\begin{figure*}[!]
\centering
\includegraphics[scale=0.75]{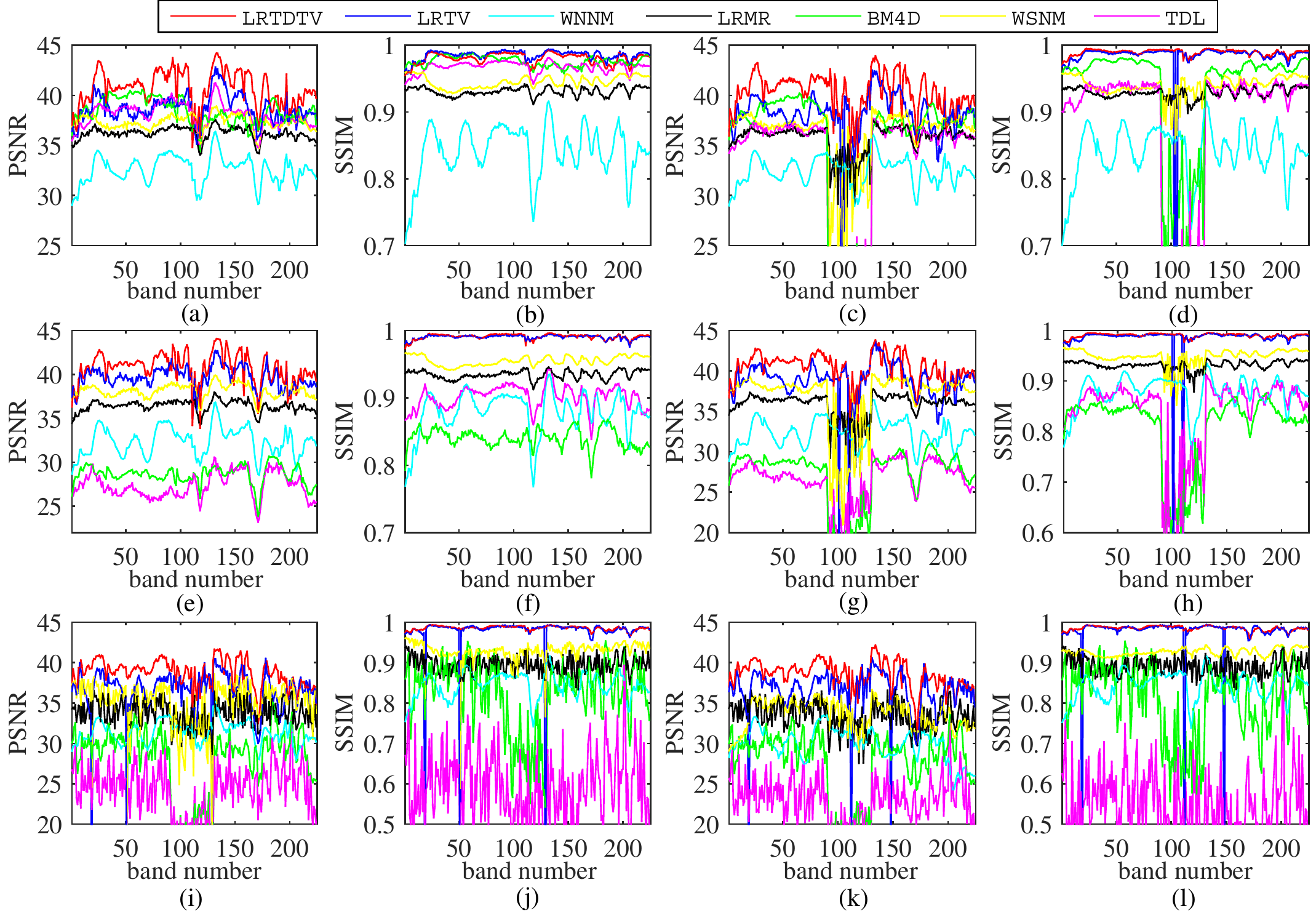}
\caption{Detailed quantitative evaluation of different methods for each band : (a-b) Case 1), (c-d) Case 2), (e-f) Case 3), (g-h) Case 4), (i)-j) Case 5), (k-l) Case 6). }
\label{fig_spect}
\end{figure*}

\begin{figure*}[!]
\centering
\includegraphics[scale=0.63]{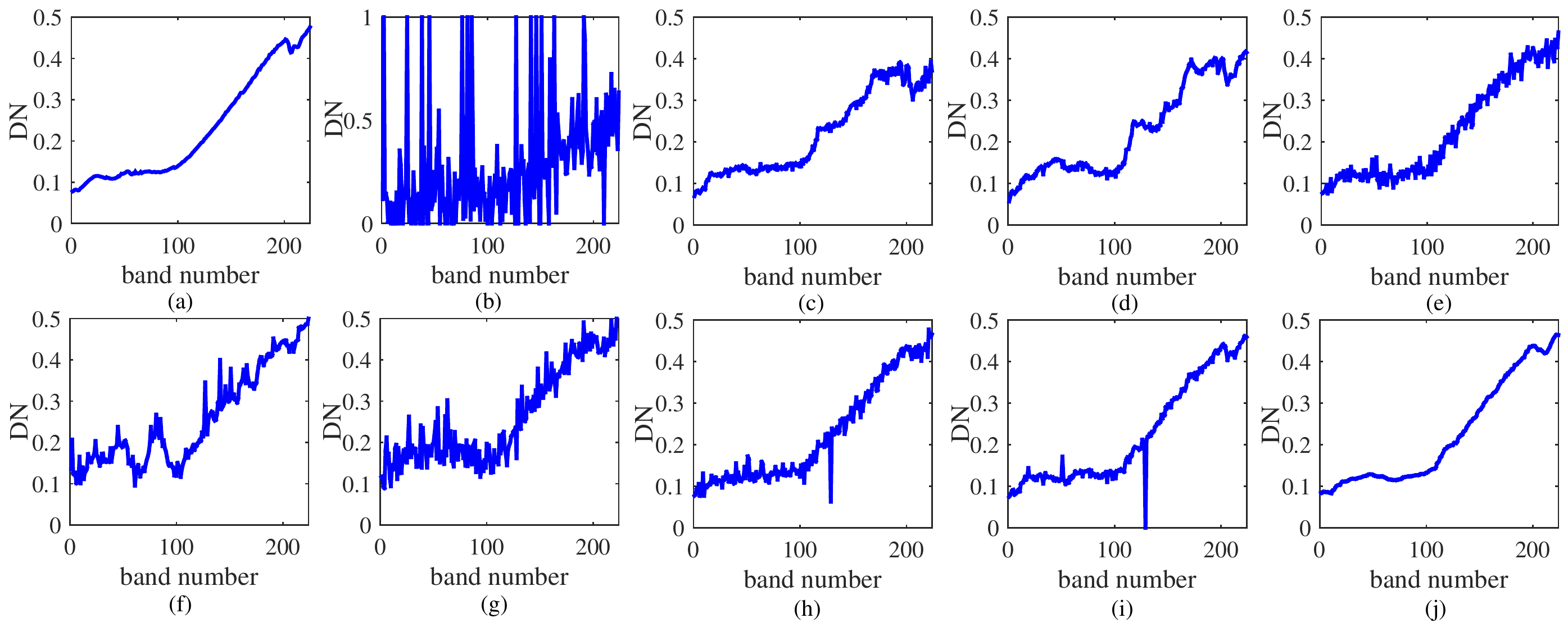}
\caption{Spectral signatures curve estimated by all the compared methods: (a) the original spectral signatures of pixel (20,20), (b) the noise spectral signatures, (c) NNM, (d) WNNM, (e) LRMR, (f) BM4D, (g) TDL, (h) WSNM, (i) LRTV, (j) LRTDTV. }
\end{figure*}  

To further compare the performances of all the restoration methods, it would be necessary to show the spectral signatures before and after restoration. As such, Fig. 8 shows one simple example namely the spectral signatures of pixel (20, 20) in case 5). Combining with the ERGAS values in Table II, it can be clearly seen that the proposed LRTDTV method produces the best spectral signature among all the compared methods.

\subsection{Real Data Experiments}
Two real-world HSI data sets were used in our experiments, i.e., the Hyperspectral Digital Imagery Collection Experiment (HYDICE) urban data set \cite{Urban} and the Airborne Visible/Infrared Imaging Spectrometer (AVIRIS) Indian Pines data set \cite{Indian}. Before conducting the restoration process, the gray values of each HSI band were normalized to [0, 1]. We have implemented the proposed two models (\ref{main_model}) and (\ref{app_model}), and indeed found that  they performed similar in these real data experiments. Therefore, we only present the restoration results of the approximation model (\ref{app_model}) for visual comparison. 
\begin{figure}[t!]
\centering
\includegraphics[width=5cm,height=5cm]{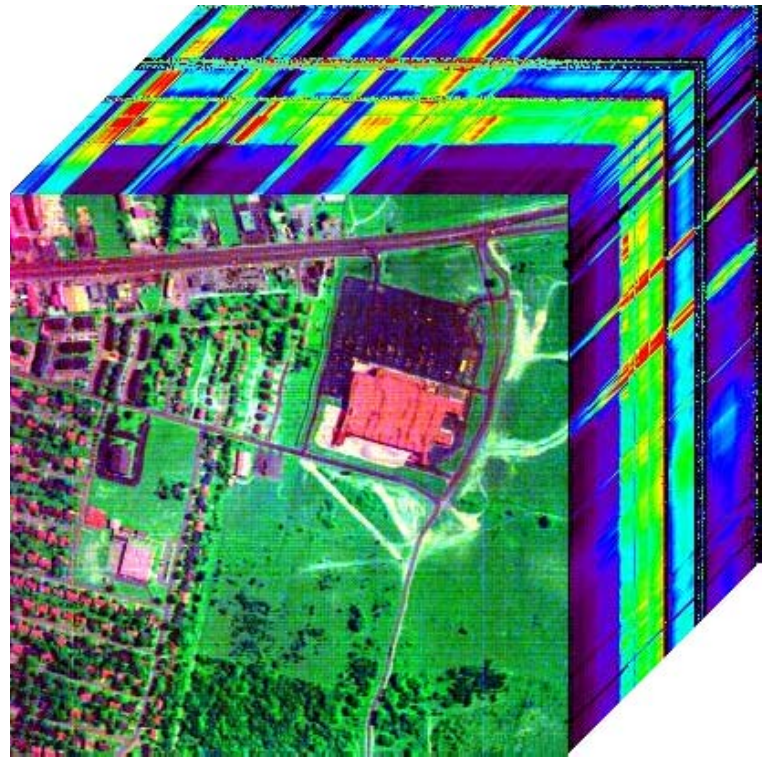}
\caption{HYDICE urban data set used in first real data experiment.}
\label{fig_urban}
\end{figure}  

 1) \textbf{HYDICE Urban Data Set}: The size of original image is $307\times 307\times 210$. As shown in Fig. 9, the full urban image is polluted by stripes, deadlines, the atmosphere, water absorption, and other unknown noise. In this experiment, NNM, WNNM, BM4D, and TDL were implemented using the default values of parameters. For other competing methods, we manually adjusted their parameters to give best visual results. For instance, the two parameters of LRMR, i.e., $r$ and $k$,  were set as 3 and 9500, respectively. 
 
 
\begin{figure*}[!]
\centering
\includegraphics[scale=0.68]{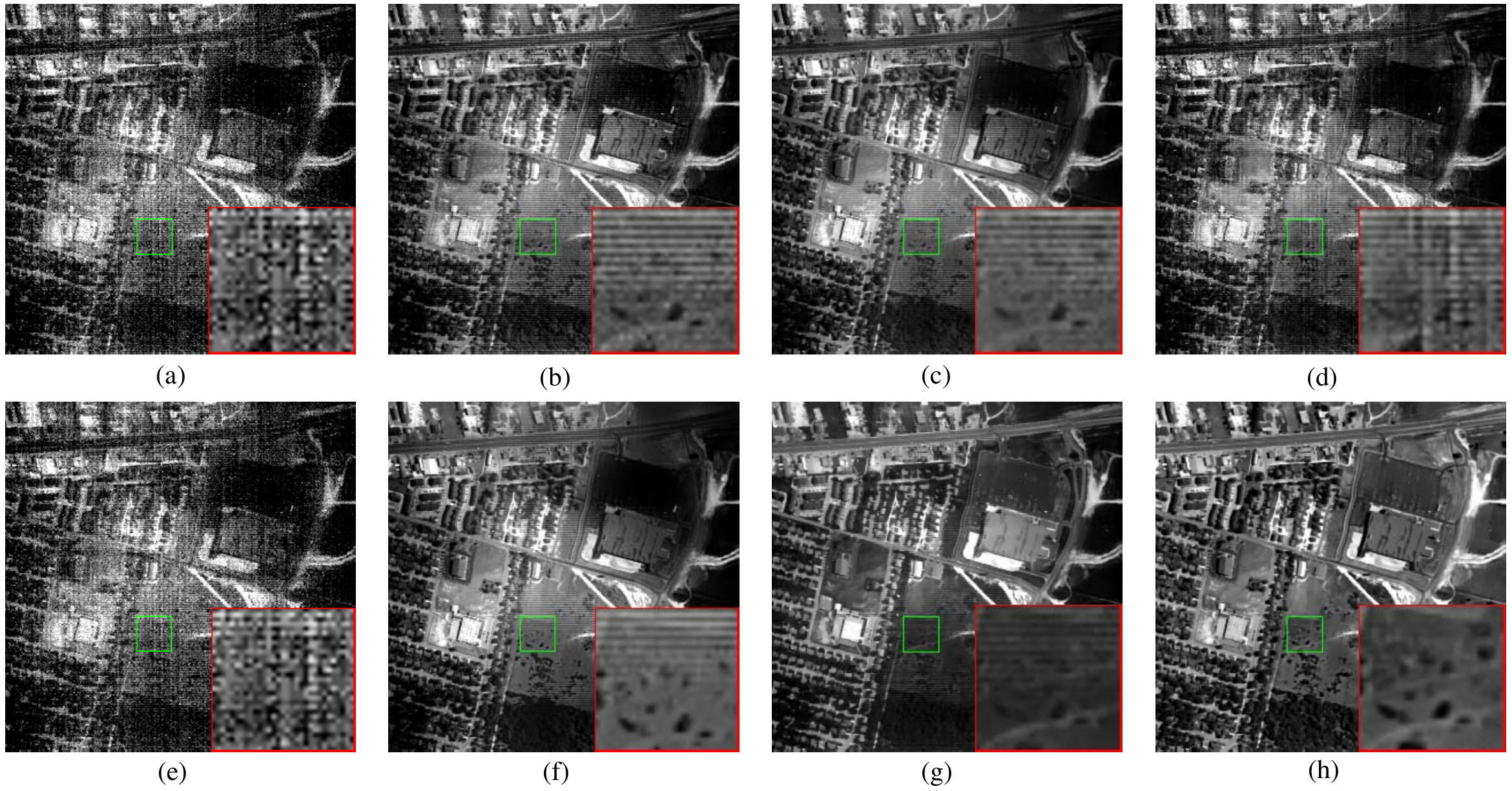}
\caption{Denoised  results by different methods: (a) the original band $109$, (b) WNNM, (c) LRMR, (d) BM4D, (e) TDL, (f) WSNM, (g) LRTV, (h) LRTDTV.}
\label{fig_U109}
\end{figure*}  

\begin{figure*}
\centering
\includegraphics[scale=0.68]{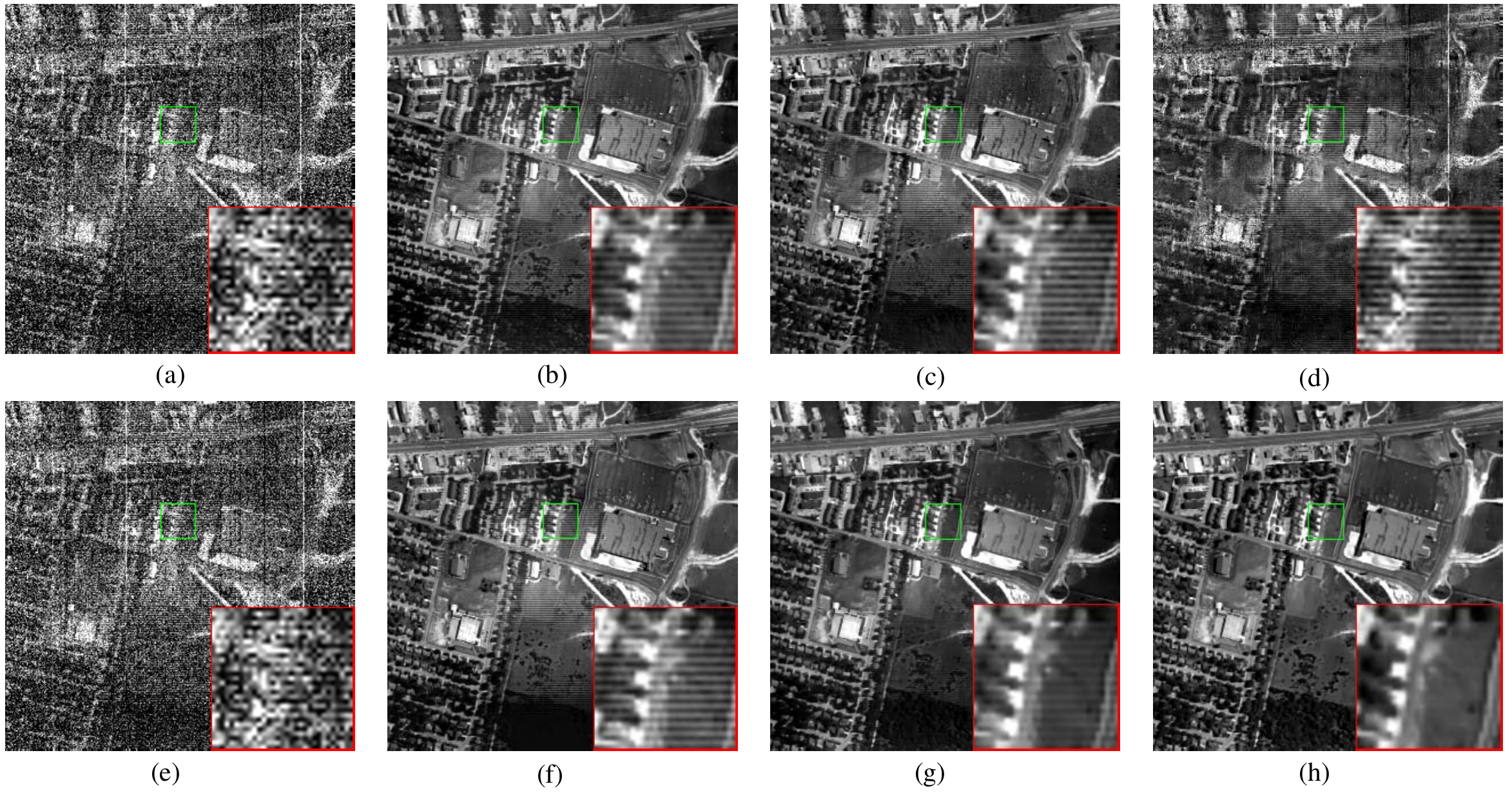}
\caption{Denoised results by different methods: (a) the original band $207$, (b) WNNM, (c) LRMR, (d) BM4D, (e) TDL, (f) WSNM, (g) LRTV, (h) LRTDTV.}
\label{fig_U207}
\end{figure*}  

In this set of experiments, we observed that NNM and WNNM have similar performance, while WNNM performed slightly better than NNM. Therefore, for simplicity,  we don't show the visual  restoration results of NNM in the following.

Figs. 10 and 11 show the restorations of  band 109 and band 207, respectively. It is not hard to see that several low-rank matrix methods including NNM, WNNM, LRMR and WSNM cannot effectively remove the stripes. In addition, some structure of objects are not clearly differentiated. This is mainly because of the fact that stripes and deadlines exist in the same place from band 104 to 110, and exist in the same place from band 199 to 210. That is, in the low rank and sparse decomposition, the stripes are more likely regarded as the low-rank part, which is assumed to be the clean image. As for BM4D, though it is a tensor filtering based method, it mainly assumes that the noise is subject to Gaussian noise, so once the HSIs suffer from heavy sparse noise, or some structure noise including stripes, this method cannot perform well. As for TDL, it loses its effectiveness when HSIs suffer from heavy mixture corrupted noise since it was also designed for removing Gaussian noise. Similar to before, since LRTV uses the spatial TV regularization to exploit the spatial information, it can remove this kind of stripes and preserve the edge information to a certain extent. By combining the spectral TV and spatial TV into a unified SSTV regularization, and utilizing the low-rank Tucker decomposition to encode the global spectral-and-spatial correlation, our proposed LRTDTV method can better remove the complex mixed-noise and preserve the spatial texture information compared with other competing methods. 

\begin{figure*}[!]
\centering
\includegraphics[scale=0.65]{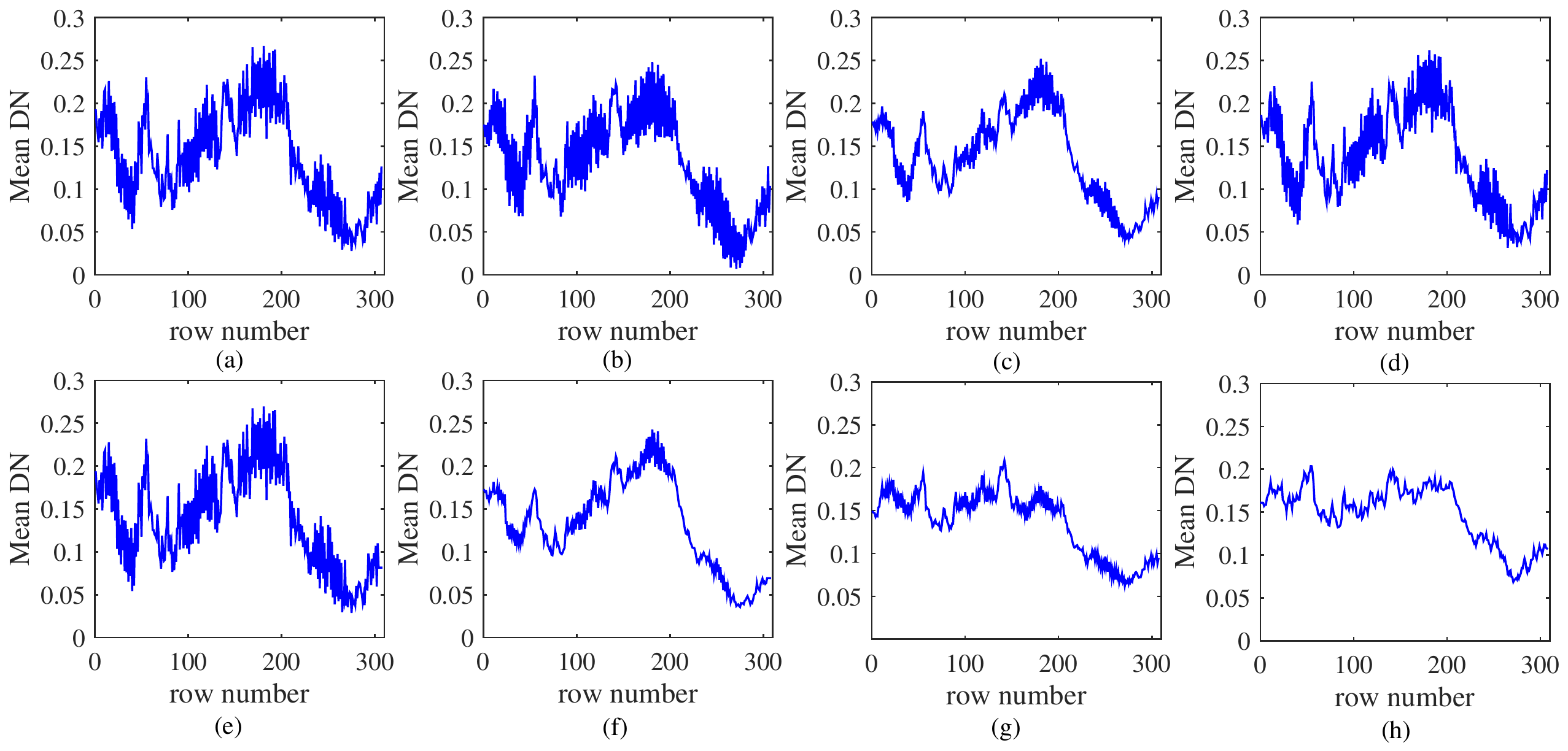}
\caption{Spectral signatures curve estimated by different methods: (a) the original band $109$, (b) WNNM, (c) LRMR, (d) BM4D, (e) TDL,  (f) WSNM, (g) LRTV, (h) LRTDTV. }
\label{fig_U109_mhd}
\end{figure*} 

We then show in Fig. 12 that the horizontal mean profiles of band 109 before and after restoration. As shown in Fig 12(a), due to the existence of mixed-noise, especially stripes and deadlines, there are rapid fluctuations in the curve before processing the restoration. After restoration, the fluctuations are more or less suppressed by all the methods. Here, one can see that the curve of proposed LRTDTV method is most stable, which is in accordance with the visual results presented in Fig. 10.

2) \textbf{AVIRIS Indian Pines Data Set}: This data set was acquired by the NASA AVIRIS instrument over the Indian Pines test site in Northwestern Indiana in 1992, and it has $145\times 145$ pixels and $220$ bands. The full urban image showed in Fig. 13 is mainly corrupted by deadlines, atmosphere, water absorption, and others. Similar to HYDICE data experiment, we implemented  WNNM, BM4D and TDL using the default parameters, and the rest using the fine-tuned parameters. 

 
\begin{figure}[!]
\centering
\includegraphics[scale=0.7]{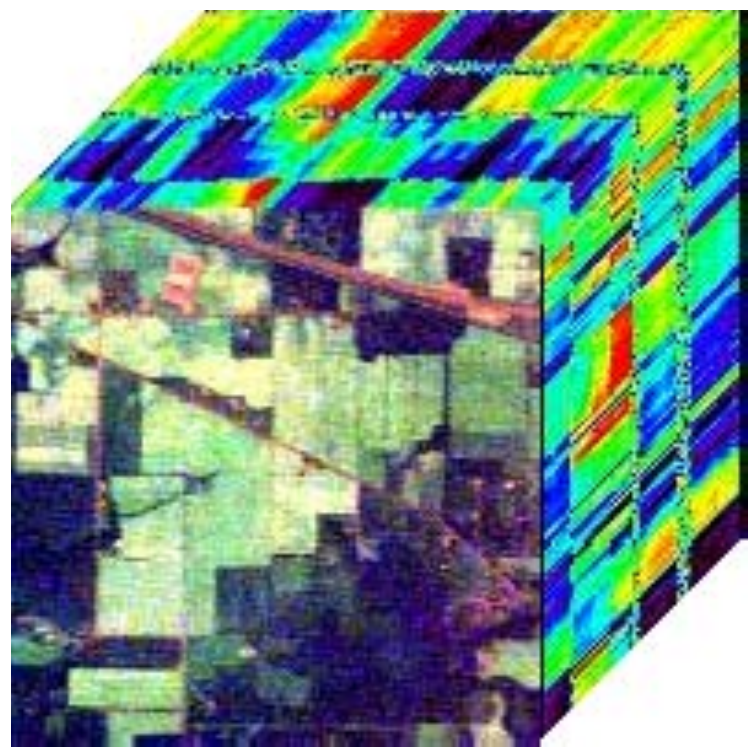}
\caption{AVIRIS Indian Pines data set used in second real data experiment. }
\label{fig_X93AV3C}
\end{figure}  


We selected two typical bands, namely band 108 and band 220, to present the performance of all the compared methods in Figs. 14 and 15. Since the TDL method completely failed to restore these two band, we excluded it  in this visual comparison. Here, it is easy to observe that low-rank matrix methods including NNM, WNNM, LRMR and WSNM can more or less remove some noise, but when the noise is heavy, such methods lose their utility and even occur some degradation of the gray value.  See Fig. 14(f) for an example of WSNM. As for BM4D, it would over-smooth the image and distort the structure as shown in Fig. 14(e),  and thus fails to give satisfactory results. Compared to the aforementioned non-TV-regularized methods, LRTV and the proposed LRTDTV can remove lots of noise, but LRTV cannot preserve edges and local detail information as well as LRTDTV. More detailed visual comparison results can be seen in such red boxes. In a word, for this data set, the proposed LRTDTV can still get best performance for removing such heavy mixed-noise. 

\begin{figure*}[!]
\centering
\includegraphics[scale=0.675]{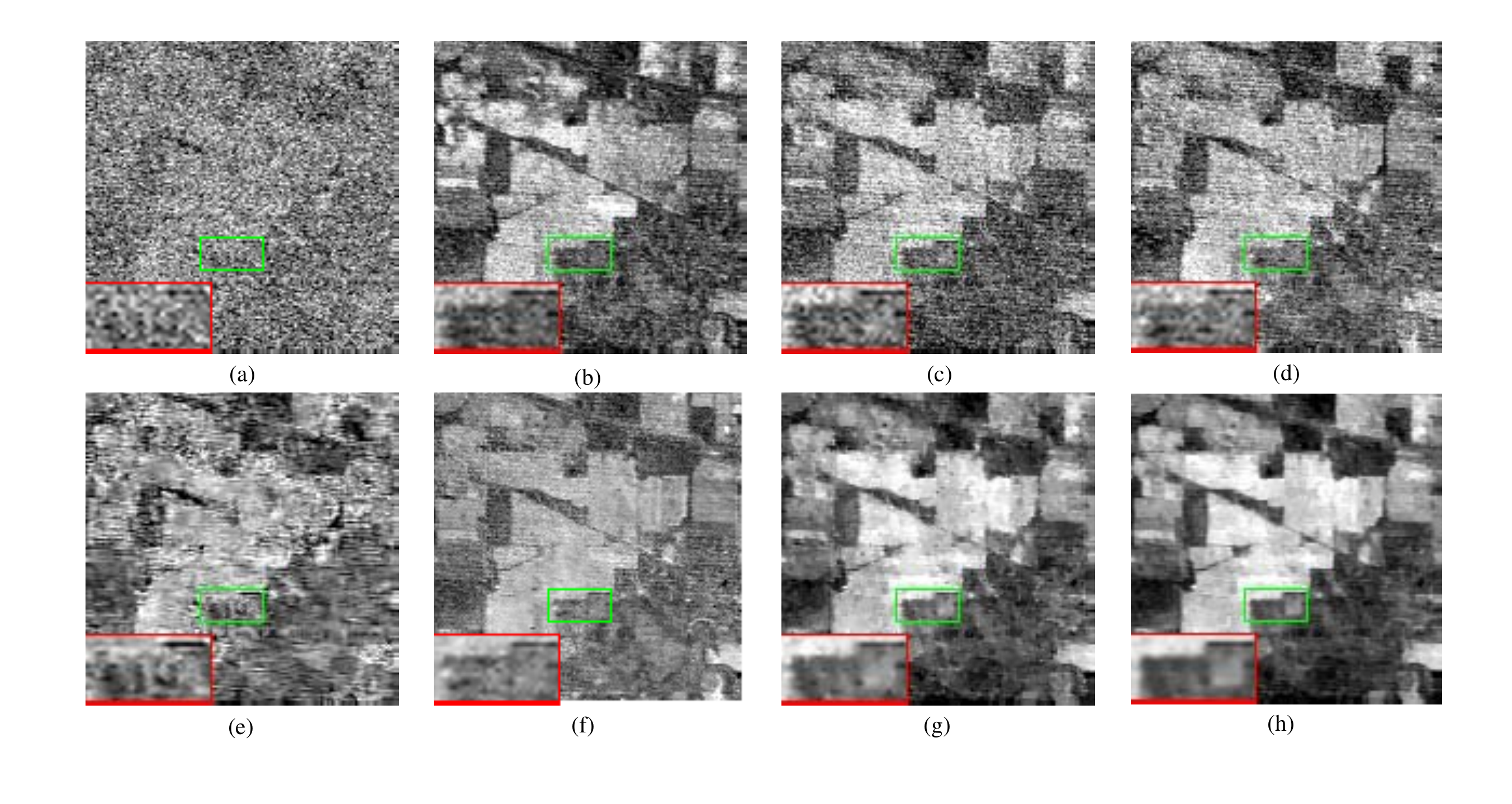}
\caption{Denoised  results by different methods: (a) the original band $108$, (b) NNM, (c) WNNM, (d) LRMR, (e) BM4D, (f) WSNM, (g) LRTV, (h) LRTDTV}
\label{fig_X108}
\end{figure*}  

\begin{figure*}
\centering
\includegraphics[scale=0.675]{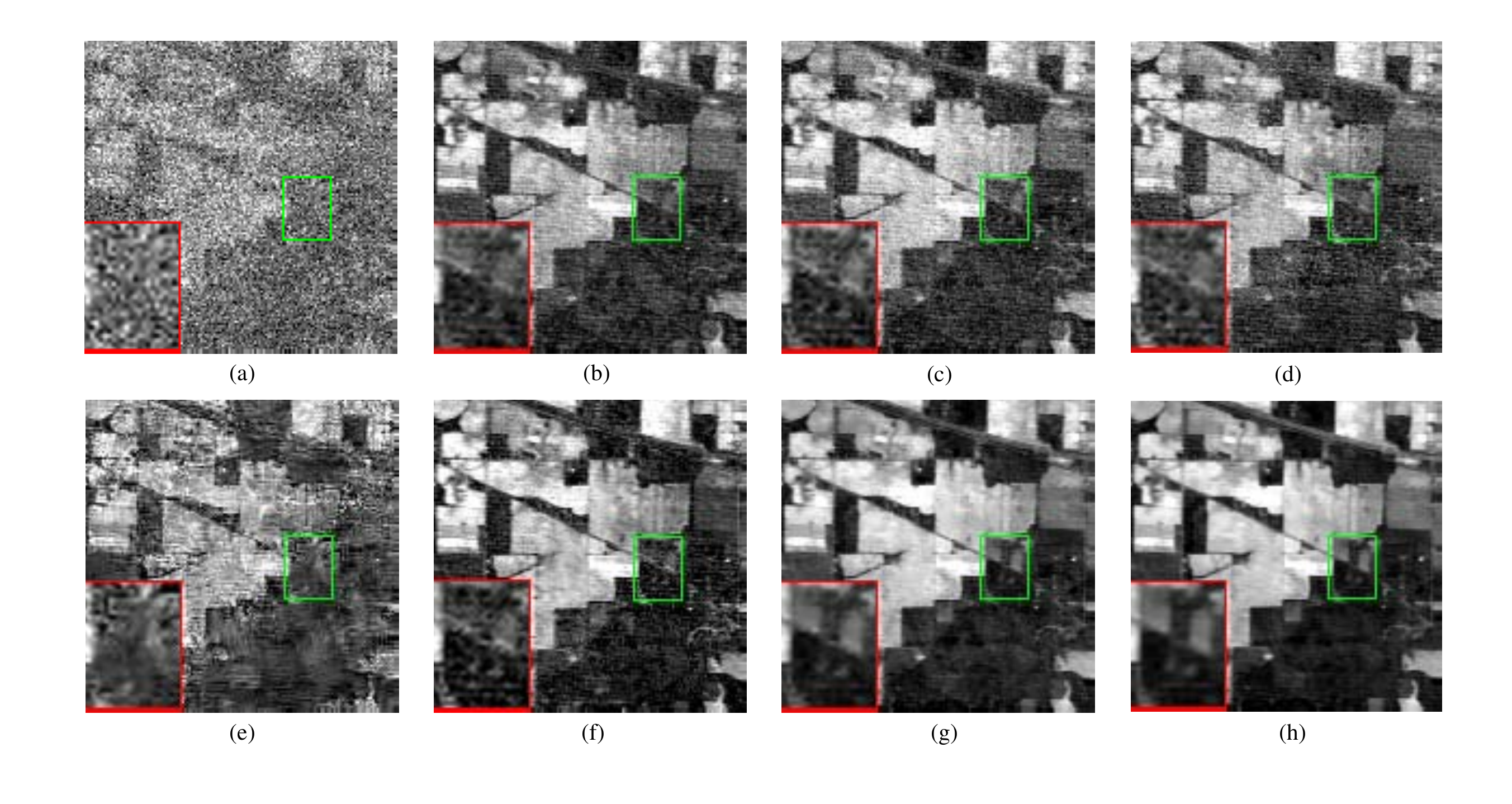}
\caption{Denoised  results by different methods: (a) the original band $220$, (b) NNM, (c) WNNM, (c) LRMR, (e) BM4D, (f) WSNM, (g) LRTV, (h) LRTDTV}
\label{fig_X220}
\end{figure*}  

We also present the vertical mean profiles of band 108 before and after restoration in Fig. 16.  Similar to Fig. 12, it can again be clearly observed that LRTDTV gives the best curve among all the restored vertical mean profiles.

To sum up, extensive experiments on both simulated and real data demonstrate the clear superiority of LRTDTV over other popular methods,  because more useful structures are exploited.

\begin{figure*}
\centering
\includegraphics[scale=0.65]{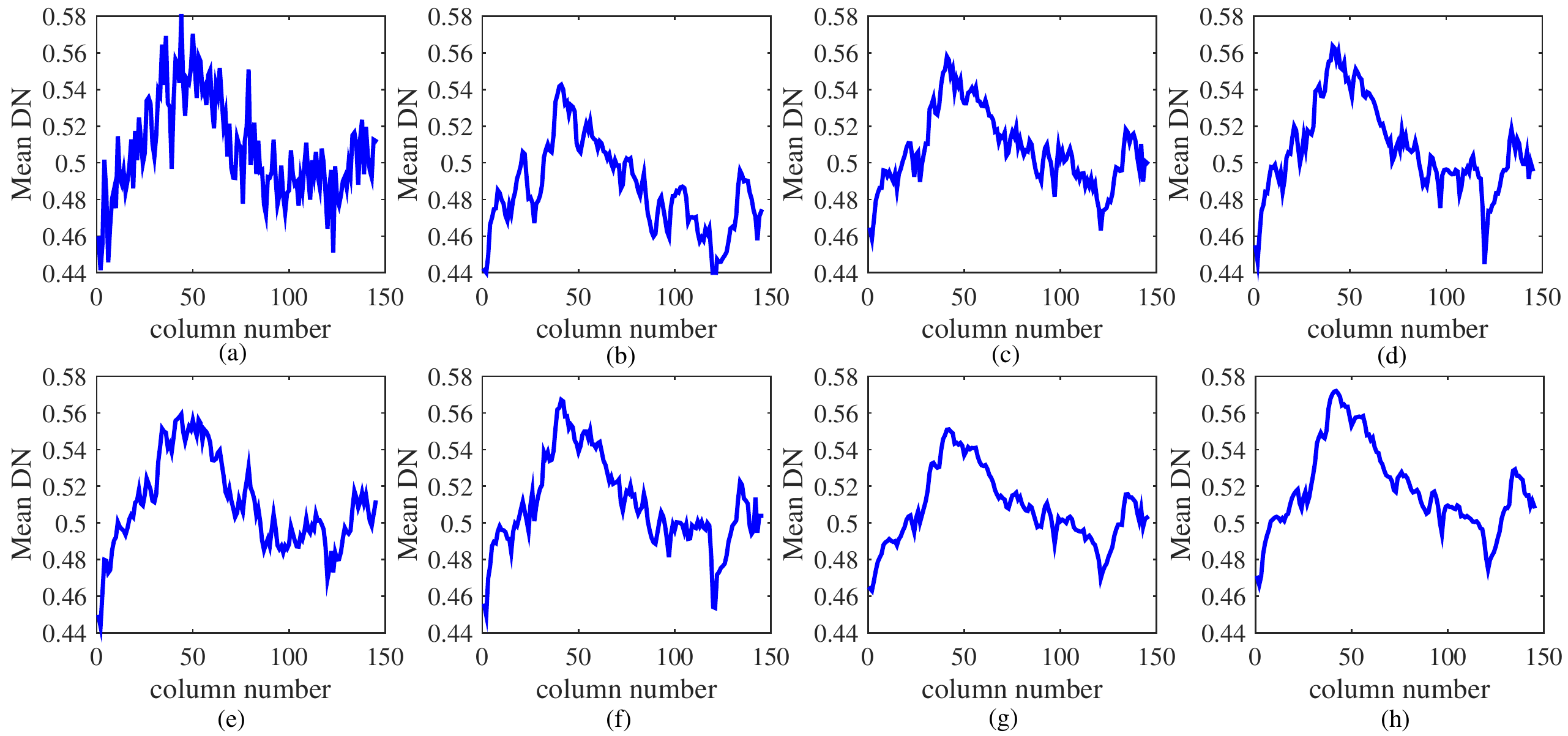}
\caption{Spectral signatures curve estimated by all the compared methods: (a) the original band $108$, (b) NNM, (c) WNNM, (d) LRMR, (e)BM4D, (f) WSNM, (g) LRTV, (h) LRTDTV. }
\label{fig_X162_mhd}
\end{figure*}

\subsection{Discussion}
 Basically, we introduced two LRTDTV models, one is the general model (\ref{main_model}), the other is the approximate model (\ref{app_model}). When the noise HSI was corrupted  by very heavy Gaussian noise, the general model (\ref{main_model}) would perform better than the approximate model (\ref{app_model}) because of the use of the additional Frobenius norm term for modeling the Gaussian noise. In fact, the noise of real HSI is usually a mixture of several types of noise, and compared to other noise terms, the extent of Gaussian noise was not severely. Thus, considering the fact that the TV regularization has the ability of removing Gaussian noise, we would like to use the approximate model (\ref{app_model}) for simplicity.

 In the LRTDTV model, there exist several parameters need to be carefully identified. Specifically, for the Frobenius term regularization parameter $\beta$, we set it as the reciprocal to the variance of Gaussian noise in general model, just as WNNM did; For the TV regularization parameter $\tau$ in two models, similar to many other works, it can be fixed as constant 1. Then, we need to consider how to select the $\ell_1$ term regularization parameter $\lambda$, the rank $\mathbf{r}_i$s among three dimensions of HSI cube, and the weights of SSTV. In the following, we shall provide the reasons for choosing such parameters in our experiments. Additionally, we present an empirical analysis of convergence of our proposed LRTDTV solver. All the results were based on the simulated data experiment in Case 5).


\begin{figure}[!]
\centering
\includegraphics[scale=0.48]{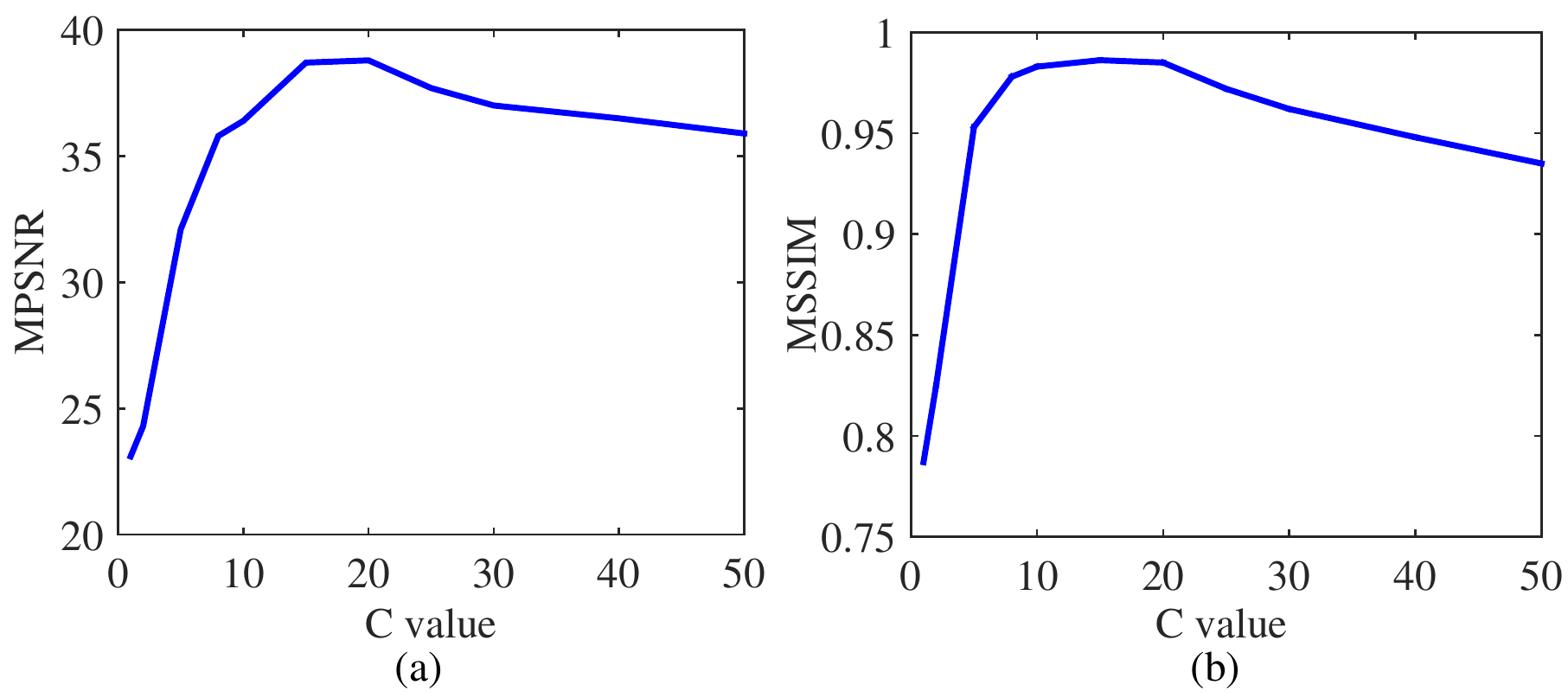}
\caption{Sensitivity analysis of parameter (C from 1 to 50 with $λ =\frac{100*C}{\sqrt{MN}}$), (a) Change in the MPSNR value, (b) Change in the MSSIM value. }
\label{fig_C_values}
\end{figure} 

1) \textbf{Sensitivity Analysis of Parameter} $\mathbf{\lambda}$: It is easy to see that $\lambda$ is the parameter used to restrict the sparsity of the sparse noise. As stated in the RPCA model\cite{Candes2011rpca}, the sparsity regularization parameter was set to $\lambda =1/\sqrt{MN}$, which was good enough to guarantee the existence of an optimal solution. Owing to bring the new TV penalty, our model is different from  RPCA, we thus set $\lambda$ as $\lambda=100\times C/\sqrt{MN}$ where $C$ is a tuning parameter. Fig. 17 shows the restoration results as $C$ varied in the set $\{1,2,5,8,10,15,20,25,30,40,50\}$. It can be clearly observed from this figure that the results of the LRTV solver are relatively stable in terms of both MPSNR and MSSIM values, with the value of  $C$ was changed from 10 to 25. Therefore, we suggest the use of $\lambda=10$ in all the simulated data experiments.

\begin{figure}[!]
\centering
\includegraphics[scale=0.48]{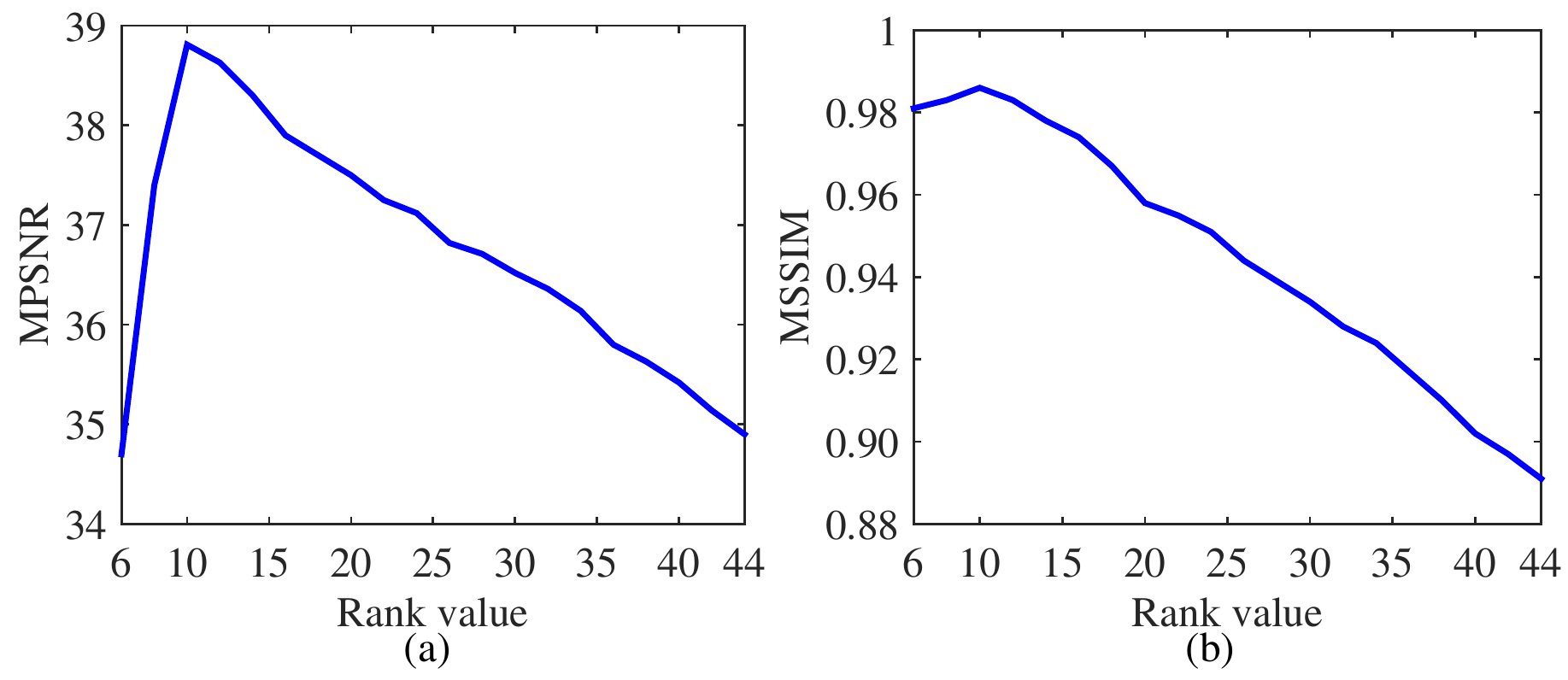}
\caption{Sensitivity analysis of the spectral rank constraint (spectral mode rank values from 6 to 44). (a) Change in the MPSNR value, (b) Change in the MSSIM value. }
\label{fig_rank_values}
\end{figure} 

2) \textbf{Effectiveness of the Rank Constraint}: In the LRTDTV optimization, we adopted the Tucker decomposition to encode the low-rank prior. So we should give the estimated ranks along the three modes before running the algorihm. For two spatial modes, we empirical adopted the eighty percent of size to ultilize the low-rank property adequately. Fig. 18 presents the MPSNR and MSSIM values of the LRTDTV solver with different rank-constrained values of spectral mode. It can be easily observed that the MPSNR and MSSIM values first increase and then decrease with the growth of the estimated rank value. This inspires us to use HSI subspace estimation method (e.g., HySime \cite{bioucas2008}) to estimated the range of values for the rank, and then choose the best desired rank value among candidates for spectral mode. As such, in all our simulation experiments, the value of spectral rank is set to 10.

\begin{figure}[!]
\centering
\includegraphics[scale=0.48]{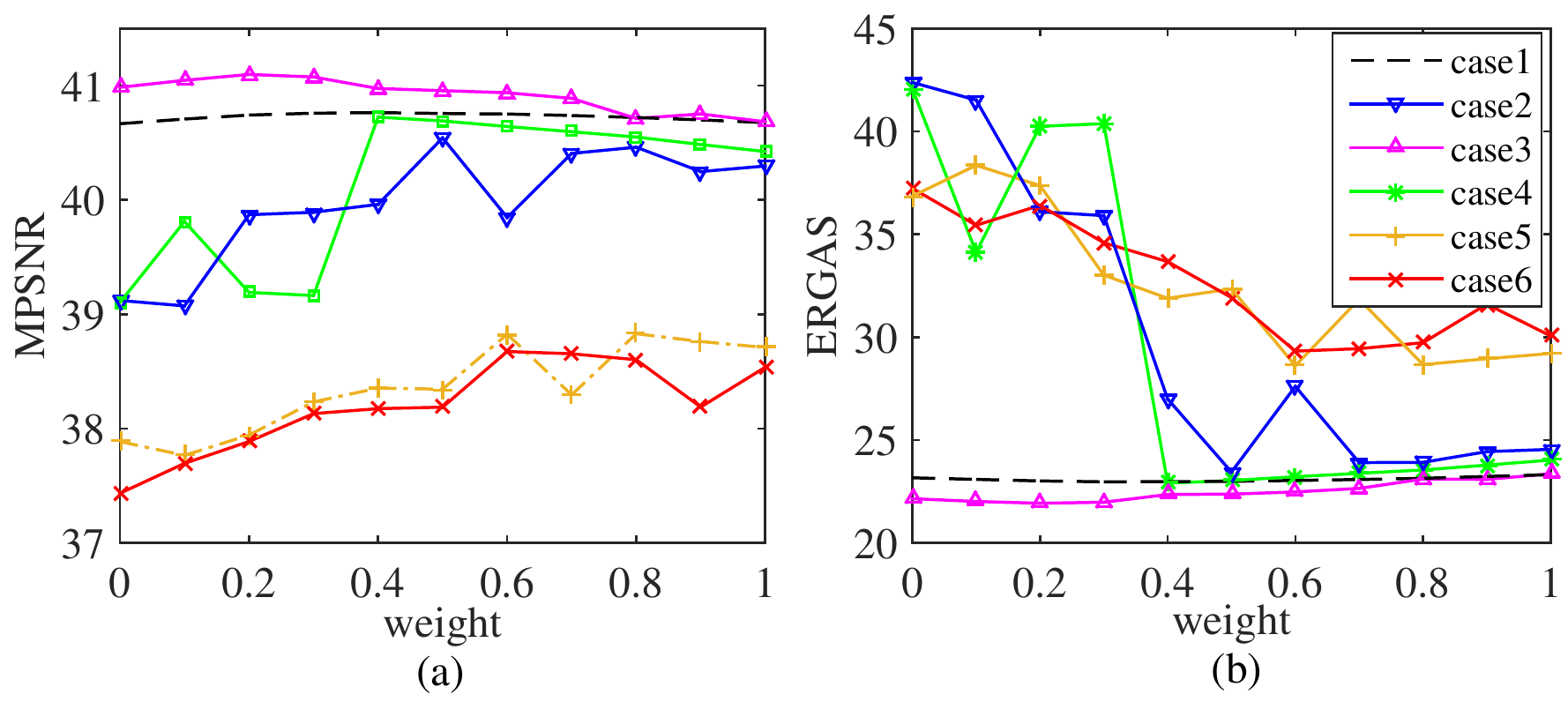}
\caption{The performance with different weights of SSTV. (a) Change in the MPSNR value, (b) Change in the ERGAS value. }
\label{fig_spect}
\end{figure}  

3) \textbf{Effectiveness of the weight of SSTV}: For hyperspectral image, the spectral characteristics is a very important feature, which can be encoded by spectral TV prior. This prior can help us to remove the noise, especially for some structure noise whose distribution was different on adjacent bands as observed. Then, based on the statistical analysis of the TV value along three modes shown in Fig. 4, the weight of spatial TV is default as 1, and we only change the weight of SSTV along the spectral mode from 0 to 1. The influence of the weight of spectral TV is presented in Fig. 19, which clearly shows the benefit of tuning the weight of SSTV. 
\begin{figure}[!]
\centering
\includegraphics[scale=0.48]{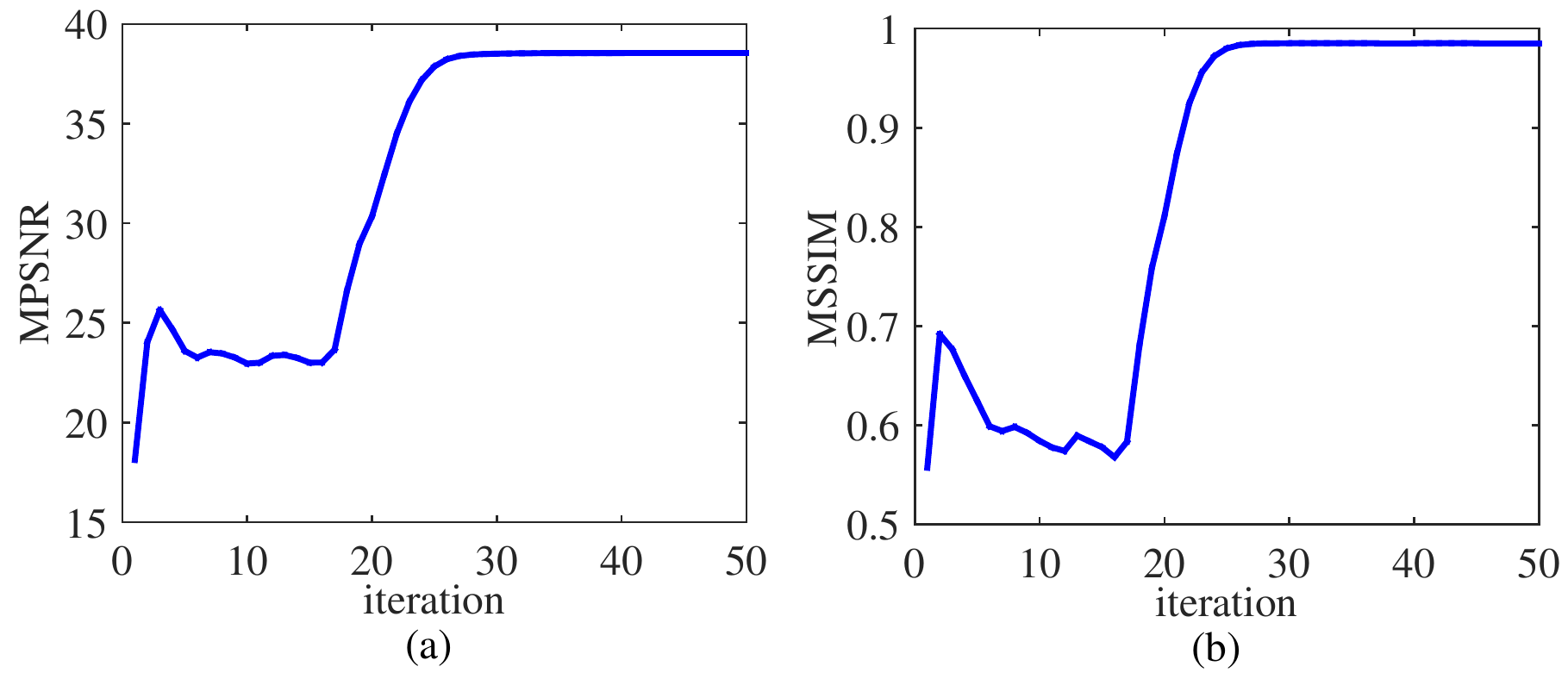}
\caption{MPSNR and MSSIM value versus the iteration number of LRTDTV. (a) Change in the MPSNR value, (b) Change in the MSSIM value. }
\label{fig_ieration}
\end{figure} 

4) \textbf{Convergence of the LRTDTV Solver}: Fig. 20 presents the MPSNR and MSSIM gains versus the iteration number of the LRTDTV solver. Here, we can observe that, as the number of iteration increasing to a relatively large value, the relative changes of MPSNR and MSSIM converge to zero. This clearly illustrates the convergent behavior of the proposed method, which further encourages us to use it for more practical situations.

\section{Conclusion}
In this paper,  we have proposed a novel tensor-based  approach for removing mixed noise in HSIs. Specifically, the low rank tensor Tucker decomposition is utilized to describe the global spatial-and-spectral correlation among all HSI bands, and a spatial-spectral total variation (SSTV) regularization is applied to characterize the piecewise smooth structure in both spatial and spectral domain of HSIs. Besides the commonly-used $\ell_1$ norm to detect the sparse noise, we have also considered an additional Frobenius norm term for modeling the heavy Gaussian noise that may exist in some practical situations. Though the resulting nonconvex tensor optimization problem seems to be difficult to solve, we have designed an efficient algorithm based on the augmented Lagrange multiplier method.  A series of simulated and real data experiments have been conducted to demonstrate the superior performance of the proposed method over some popular methods in terms of both the quantitative evaluation and the visual comparison.

In the future, we are interested in conducting the following studies.  Firstly, incorporate the noise modeling idea of \cite{chen2017tensor} into our SSTV regularized low rank tensor decomposition framework to further enhance its capability for removing more complex noise in some real-word scenarios.  Secondly, extend the deep learning ideas of \cite{gregor2010learning, Sprechmann2015learning} to design a  deep tensor architecture to learn the multilinear structure of clean HSI and identify the noise structures of observered HSI. Finally, design a new SSTV regularizer to exploit the gradient similarity among all HSI bands to further characterize the smooth structure, and as a result, improve the noise removal capability of the proposed procedure. 

\section*{Acknowledgment}

This work was supported by the National Natural Science Foundation of China (Grant Nos. 11501440, 61603292, 61673015, 61373114).







%

\bibliographystyle{IEEEtran}
\bibliography{Spars_new}


\end{document}